\documentclass{article}

\usepackage{microtype}
\usepackage{graphicx}
\usepackage{subcaption}
\usepackage{booktabs} %

\usepackage{hyperref}

\usepackage[export]{adjustbox} %
\usepackage{multirow}
\usepackage{colortbl} %
\usepackage{xcolor}   %
\usepackage[utf8]{inputenc}
\usepackage{tcolorbox}
\usepackage{tabularx}
\usepackage{ragged2e}
\usepackage{array}
\usepackage{pifont}
\usepackage{float}
\usepackage{placeins}
\usepackage{marvosym}
\usepackage{amsmath}
\usepackage[ruled,vlined,algo2e]{algorithm2e}
\usepackage{makecell}
\usepackage{tcolorbox}
\tcbuselibrary{skins,breakable}
\usepackage[table,dvipsnames,svgnames]{xcolor}
\definecolor{dimgray}{rgb}{0.41, 0.41, 0.41}
\usepackage{listings}
\usepackage[preprint]{icml2026}

\makeatletter
\renewcommand{\printAffiliationsAndNotice}[1]{%
  \global\icml@noticeprintedtrue%
}
\makeatother

\usepackage{amsmath}
\usepackage{amssymb}
\usepackage{mathtools}
\usepackage{amsthm}

\usepackage[capitalize,noabbrev]{cleveref}

\theoremstyle{plain}

\theoremstyle{definition}

\theoremstyle{remark}

\usepackage[textsize=tiny]{todonotes}

\icmltitlerunning{GLEAM: Learning to Match and Explain in Cross-View Geo-Localization}

\begin{document}

\twocolumn[
  \icmltitle{GLEAM: Learning to Match and Explain in Cross-View Geo-Localization}

\icmlsetsymbol{equal}{*}

\begin{icmlauthorlist}
  \icmlauthor{Xudong Lu\textsuperscript{*}}{cuhk}
  \icmlauthor{Zhi Zheng\textsuperscript{*}}{cuhk}
  \icmlauthor{Yi Wan}{whu}
  \icmlauthor{Yongxiang Yao}{whu}
  \icmlauthor{Annan Wang}{ntu}
  \icmlauthor{Renrui Zhang}{cuhk}
  \icmlauthor{\textbf{Panwang Xia}}{whu}
  \icmlauthor{\textbf{Qiong Wu}}{whu}
  \icmlauthor{\textbf{Qingyun Li}}{hit}
  \icmlauthor{\textbf{Weifeng Lin}}{cuhk}
  \icmlauthor{\textbf{Xiangyu Zhao}}{sjtu}
  \icmlauthor{\textbf{Peifeng Ma}}{cuhk}
  \icmlauthor{\textbf{Xue Yang\textsuperscript{\textrm{\Letter}}}}{sjtu}
  \icmlauthor{\textbf{Hongsheng Li\textsuperscript{\textrm{\Letter}}}}{cuhk}
\end{icmlauthorlist}

\vspace{0.2em}
\begin{center}
{%
\textsuperscript{1}The Chinese University of Hong Kong\quad
\textsuperscript{2}Wuhan University\quad
\textsuperscript{3}Nanyang Technological University\\
\textsuperscript{4}Harbin Institute of Technology\quad
\textsuperscript{5}Shanghai Jiao Tong University
\\ \textsuperscript{*}Equal contribution
\quad $^\textrm{\Letter}$\ Corresponding author\par
}
\end{center}
\vspace{0.15em}
\begin{center}
{%
{\ttfamily\detokenize{{luxudong@link, hsli@ee}.cuhk.edu.hk}}\quad
{\ttfamily\detokenize{zhizheng@cuhk.edu.hk}}\\
{\ttfamily\detokenize{yangxue-2019-sjtu@sjtu.edu.cn}}
}
\end{center}

  \icmlkeywords{Machine Learning, ICML}

  \vskip 0.3in
]

\printAffiliationsAndNotice{}  %

\begin{abstract}
Cross-View Geo-Localization (CVGL) focuses on identifying correspondences between images captured from distinct perspectives of the same geographical location. However, existing CVGL approaches are typically restricted to a single view or modality, and their direct visual matching strategy lacks interpretability: they only determine whether two images correspond, without explaining the rationale behind the match. In this paper, we present GLEAM-C, a foundational CVGL model that unifies multiple views and modalities by aligning them exclusively with satellite imagery. Our framework improves training efficiency through optimized implementation and achieves accuracy comparable to prior modality-specific CVGL models via a novel two-phase training strategy. To address interpretability, we further propose GLEAM-X, a novel task that combines cross-view correspondence prediction with explainable reasoning enabled by multimodal large language models (MLLMs). We construct a bilingual benchmark using commercial MLLMs to generate training and testing data, and refine the test set through rigorous human revision for systematic evaluation of explainable cross-view reasoning. Together, GLEAM-C and GLEAM-X form a comprehensive CVGL pipeline that integrates multi-modal, multi-view alignment with interpretable correspondence analysis, unifying accurate cross-view matching with explainable reasoning and advancing \textbf{G}eo-\textbf{L}ocalization by enabling models to better \textbf{E}xplain \textbf{A}nd \textbf{M}atch. Code and datasets used in this work will be made publicly accessible at \url{https://github.com/Lucky-Lance/GLEAM}.

\end{abstract}

{\vspace*{-1em}}
\section{Introduction}\label{sec:intro}

Cross-View Geo-Localization (CVGL) seeks to determine the geographic position of a query image by establishing correspondences with a geo-referenced database captured from alternative viewpoints, %
{especially satellite observations~\citep{chen2025multi, li2024unleashing, xia2024enhancing}. Unlike Global Navigation Satellite Systems (GNSS), which suffer from signal blockage and noise in complex environments, CVGL offers a complementary and robust vision-based localization paradigm.} This capability has broad relevance to autonomous driving~\citep{cui2003autonomous,chen2023gnss}, robotic navigation~\citep{nowak2024enhancing,semborski2024review}, unmanned aerial vehicles (UAVs) navigation~\citep{xu2024precise,suzuki2016precise,wang2024multiple}, and augmented reality devices~\citep{kamalam2022augmented,sathyanarayana2020comparison}, where reliable positioning forms the foundation for effective operation.

\begin{figure*}[t]
    \centering
    \includegraphics[width=\linewidth]{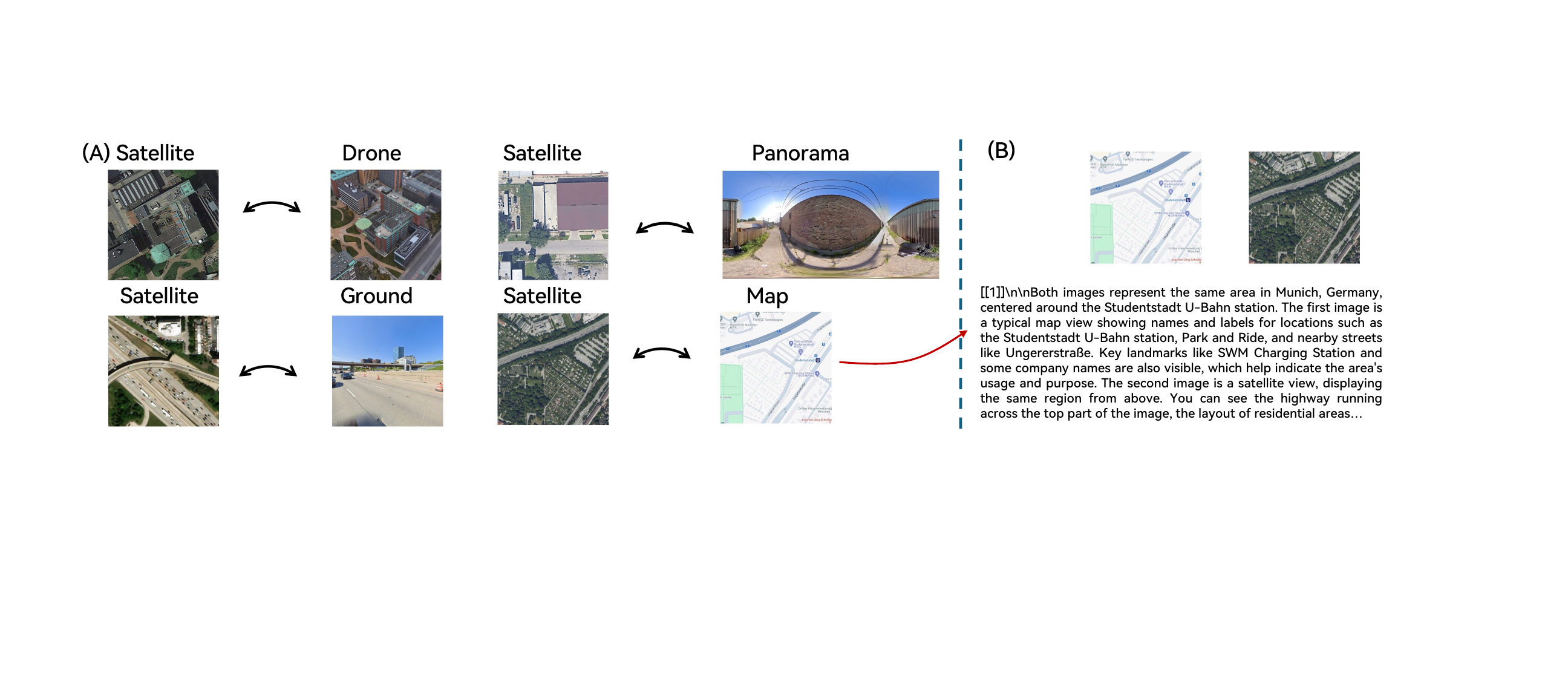}
    \caption{Overview of GLEAM-Core and GLEAM-eXplain. (A) GLEAM-C: a novel foundation CVGL model trained to align multiple views and modalities with satellite imagery, including UAV imagery, street maps, panoramic images, and ground photos. (B) GLEAM-X: a novel benchmark combining cross-view correspondence prediction with explainable reasoning. We illustrate a representative example between query (street map) images and satellite images.}
    \label{fig:overview}
\end{figure*}

Despite its practical importance, CVGL remains a technically demanding problem. The task is complicated by the appearance gap across heterogeneous viewpoints and modalities~\citep{Regmi2019BridgingTD, Ge2024MultibranchJR}. Consequently, prior studies have typically focused on a single viewpoint and modality, such as UAV imagery~\citep{zheng2020university}, street maps~\citep{Zhao2021DeepLH}, panoramic views~\citep{zhu2021vigor}, or ground photographs~\citep{wu2024cross}. While these modality-specific approaches have achieved promising results, the problem of establishing robust cross-modal alignment within a unified framework remains unresolved. A unified CVGL framework offers several advantages. First, it enables multi-platform deployment through a single inference interface, reducing system complexity, memory footprint, and initialization latency on resource-constrained platforms such as drones and vehicles. Second, it improves robustness and generalization by learning a geographically grounded representation that captures diverse structural regularities. By bridging heterogeneous views and modalities, such a system enables robust, generalizable, and efficient geo-positioning across diverse real-world scenarios.

Moreover, the dominant paradigm in CVGL has been image retrieval or binary correspondence prediction, which offers limited interpretability~\citep{Deuser2023Sample4GeoHN, Xia2024EnhancingCG}. For safety-critical applications such as navigation or disaster response, a mere correspondence score is insufficient; instead, models must provide interpretable reasoning about why two views are matched or mismatched. Recent advances in multimodal large language models (MLLMs) resolve this limitation. By jointly processing visual and textual information, MLLMs can reason about image correspondences and generate human-interpretable explanations~\citep{achiam2023gpt4, Bai2025Qwen25VLTR, Zhu2025InternVL3EA}. Integrating these reasoning capabilities into CVGL enables models not only to predict whether two images correspond but also to articulate the underlying rationale, enhancing transparency and accountability.

To address these two challenges, in this paper, we propose a comprehensive solution that combines robust multi-modal alignment with interpretable cross-view reasoning. The overview of our work is shown in Fig.~\ref{fig:overview}. First, we design GLEAM-Core (GLEAM-C), a foundational CVGL model that integrates multiple views and modalities—including UAV imagery, street maps, panoramic views, and ground photographs—by aligning them exclusively with satellite images. This unified design eliminates the need for modality-specific architectures, thereby simplifying the modeling process and improving scalability across heterogeneous data sources. In GLEAM-C, we reconstruct the conventional Data Parallel (DP) training strategy used in prior work~\citep{deuser2023sample4geo} into a Distributed Data Parallel (DDP) scheme, which yields over 5-fold improvement in training efficiency. We design a novel two-phase training strategy and evaluate GLEAM-C on both CNN- and ViT-based architectures. The results show accuracy comparable to that of modality-specific models. Second, we design GLEAM-eXplain (GLEAM-X), which novelly combines cross-view correspondence prediction with explainable reasoning. To enhance model robustness, we select 50k query images from the multi-view matching training set, each paired with a positive reference image (match) and a negative reference image (mismatch). After assigning ground-truth labels, GPT-4o and Doubao-1.5-Thinking-Vision-Pro generate explanations in both Chinese and English, clarifying the reasons for image correspondences. For evaluation, we select 504 query images from the matching test set, each paired with a positive and a negative reference image. After GPT-4o generates explanations, we perform multiple rounds of rigorous human annotation and refinement to create a high-quality test set. We fine-tune the Qwen2.5-VL-3B-Instruct MLLM on the training set, achieving higher matching accuracy than GPT-4o and Doubao-1.5, with analysis results closely aligned with human annotations. GLEAM-C and GLEAM-X can be further integrated: GLEAM-C performs the core task of CVGL by aligning query images with geo-referenced satellite imagery. GLEAM-X enhances this process by verifying and providing human-interpretable explanations for the image correspondences predicted by GLEAM-C, improving both the model's robustness and transparency. Our contributions are summarized as follows:

\begin{table*}[t]\small
    \centering
    \renewcommand{\arraystretch}{0.8}
    \caption{Statistics of training pairs. GLEAM-C primarily constructs training pairs from existing CVGL datasets. GLEAM-X balances the data distribution across datasets and then samples positive and negative pairs, providing bilingual explanation annotations.}
\vspace{-0.2em}
\resizebox{\textwidth}{!}{%
\begin{tabular}{lccccc}
\toprule
\textbf{Dataset} & \textbf{Original Pairs} & \multicolumn{2}{c}{\textbf{GLEAM-C}} & \multicolumn{2}{c}{\textbf{GLEAM-X}} \\ 
\cmidrule(lr){3-4} \cmidrule(lr){5-6}
 &  & Sampling Ratio & Training Pairs & Sampling Ratio & \begin{tabular}[c]{@{}c@{}}Training Pairs \\ (pos+neg, en+zh)\end{tabular} \\ 
\midrule
University-1652~\citep{zheng2020university} & 37,854  & 1.00 & 37,854  & 0.36 & 54,000  \\ 
VIGOR~\citep{zhu2021vigor}           & 52,609  & 1.00 & 52,609  & 0.26 & 54,000  \\ 
SetVL-480K~\citep{Wu2024CrossViewIS}      & 240,544 & 0.50 & 120,272 & 0.06 & 54,000  \\ 
MAP             & 10,208  & 4.00 & 40,832  & 1.00 & 40,832  \\ 
\midrule
\textbf{Total}  & \textbf{341,215} & — & \textbf{251,567} & — & \textbf{202,832} \\ 
\bottomrule
\end{tabular}%
}
\label{tab:sample-ratio}
\vspace{-1.1em}
\end{table*}

\textbf{1) Problem Analysis}: We provide a detailed analysis of the CVGL task, highlight the benefits of a unified framework, and identify the lack of interpretability in current solutions.

\textbf{2) GLEAM-C}: We propose GLEAM-C, a foundation model that unifies multiple views and modalities by aligning with satellite images. We design a two-phase training strategy and leverage DDP to optimize training efficiency, achieving competitive accuracy with modality-specific CVGL models.

\textbf{3) GLEAM-X}: We introduce the novel GLEAM-X benchmark, combining CVGL with explainable reasoning. We provide a high-quality bilingual training and testing dataset, along with the Qwen2.5-VL-3B-Instruct model fine-tuned on this dataset. This enables the generation of human-interpretable explanations for image correspondences, enhancing the robustness and transparency of the model.

\textbf{4) Integrated Pipeline}: GLEAM-C and GLEAM-X can be combined into an integrated CVGL pipeline. GLEAM-C performs the core task of CVGL by aligning query images with satellite imagery, while GLEAM-X enhances this process by verifying and providing human-interpretable explanations for the predicted correspondences, improving both the model’s robustness and transparency.

\section{Related Works}\label{sec:related}

\subsection{Cross-View Geo-Localization}

Cross-View Geo-Localization (CVGL) has garnered substantial attention within the research community. Most existing research, however, focuses on a single view or modality. Early efforts construct paired ground-to-aerial datasets~\citep{lin2015learning}, which later evolve into widely used benchmarks such as CVUSA~\citep{workman2015wide}, CVACT~\citep{liu2019lending}, VIGOR~\citep{zhu2021vigor}, University-1652~\citep{zheng2020university}, and DenseUAV~\citep{dai2023vision}, enabling evaluations across panoramas, UAV imagery, and related scenarios. Methodologically, handcrafted descriptors~\citep{bansal2011geo, castaldo2015semantic} soon give way to deep learning, where pre-trained CNNs and fine-tuning strategies significantly advance cross-view correspondence~\citep{krizhevsky2012imagenet, workman2015wide}. Subsequent studies introduce a variety of techniques to alleviate viewpoint discrepancies, including polar and optimal transport transformations~\citep{shi2020optimal, shi2019spatial}, region-level and latent alignment~\citep{dai2021transformer, xia2024enhancing}, and strategies that enhance scene discrimination, such as hard-negative mining~\citep{deuser2023sample4geo} and cross-dimension interactions~\citep{shen2023mccg}. More recently, unsupervised approaches explore how to exploit unlabeled data for training~\citep{li2024learning, li2024unleashing}. Despite these advancements, most studies remain modality-specific, and the development of a unified framework that accommodates diverse modalities still represents an open challenge.

\subsection{Multimodal Large Language Model}

Multimodal large language models (MLLMs) have emerged in recent years as a powerful tool for handling complex tasks by incorporating visual inputs into traditional language models~\citep{Achiam2023GPT4TR, liu2023llava, chen2024internvl, Bai2025Qwen25VLTR}. Prominent commercial models such as OpenAI’s GPT-4o~\citep{hurst2024gpt} and Google’s Gemini 2.5~\citep{comanici2025gemini} exemplify this trend by simultaneously handling text, audio, images, and video, unlocking new capabilities like real-time, human-like voice interaction and long-context video understanding. In addition to closed-source commercial MLLMs, numerous well-known open-source models, such as InternVL3~\citep{wang2025internvl3}, Qwen2.5-VL~\citep{bai2025qwen2}, MiniCPM-V 4.5~\citep{yao2024minicpm}, and LLaVA-OneVision~\citep{li2024llava}, have made significant contributions to advancing multimodal understanding. Beyond perception, recent efforts increasingly emphasize unifying multimodal understanding and generation, thereby enabling models not only to interpret visual inputs but also to synthesize new content. Representative examples include Janus-Pro~\citep{chen2025janus}, Show-o~\citep{xie2024show}, and BAGEL~\citep{deng2025bagel}, which collectively mark a transition toward general-purpose multimodal intelligence that integrates both reasoning and creativity.

\begin{figure*}
    \centering
    \includegraphics[width=0.95\linewidth]{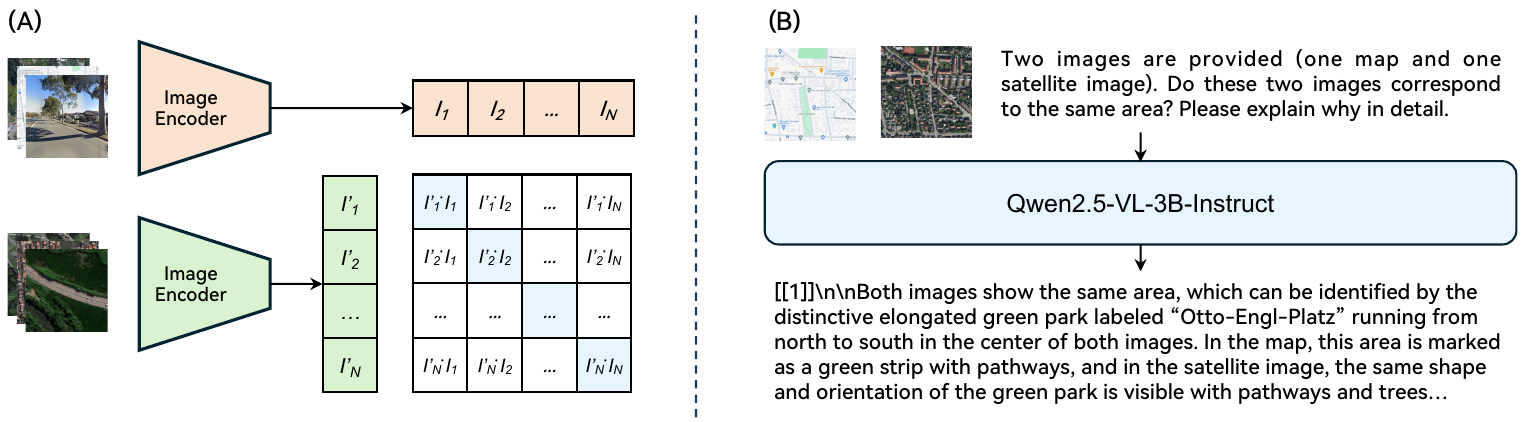}
    \caption{Method overview of GLEAM-C and GLEAM-X. (A) GLEAM-C: We design a novel two-phase contrastive learning paradigm to train the CVGL model across UAV, street map, panoramic, and ground photographs. (B) GLEAM-X: This component formulates a novel multi-image reasoning task in CVGL. The MLLM receives a query image, a reference image, and a natural language instruction. Through fine-tuning, it delivers both a matching prediction and an interpretable textual explanation.}
    \label{fig:method}
\end{figure*}

\section{GLEAM-C: A Unified CVGL Core Model}~\label{sec:gleam-c}

{\vspace*{-1.2em}}

GLEAM-C is a unified CVGL model aligning visual representations across diverse viewpoints and modalities. In this section, we provide the data composition (Sec.~\ref{sec:c-data}), model structure (Sec.~\ref{sec:c-model}), and training recipe (Sec.~\ref{sec:c-train}).

\subsection{Data Composition}\label{sec:c-data}

GLEAM-C is a foundational CVGL model that novelly aligns multiple views and modalities exclusively with satellite imagery, including UAV imagery, street maps, panoramic views, and ground photographs. To facilitate training, we utilize widely adopted datasets from prior work: University-1652~\citep{zheng2020university} for UAV imagery, VIGOR~\citep{zhu2021vigor} for panoramic views, and SetVL-480K~\citep{Wu2024CrossViewIS} for ground photographs. In addition, for street maps, we manually collect 12,761 pairs of corresponding street maps and satellite images from Google Maps, with each image resized to 512$\times$512 pixels.

The original datasets show highly imbalanced sample distributions across different views and modalities, as shown in Tab.~\ref{tab:sample-ratio}, which could introduce biases and hinder the model from learning uniformly. To mitigate this issue, we employ data sampling to balance the number of samples across all views and modalities. This ensures that the model receives sufficient and diverse training signals from each type of data, improving its robustness across heterogeneous inputs. For evaluation, we adhere to the original test datasets and protocols provided by each benchmark. The newly introduced MAP test follows the same evaluation logic as VIGOR.

\subsection{Model Structure}\label{sec:c-model}

We leverage a contrastive training strategy for GLEAM-C, where the query and reference images share the same encoder for feature extraction. The InfoNCE loss is applied to learn discriminative features across both view directions. The training framework is shown in Fig.~\ref{fig:method}A. Our implementation is based on the well-engineered Sample4Geo~\citep{deuser2023sample4geo} codebase. Notably, Sample4Geo introduces GPS-Sampling (GPS) and Dynamic Similarity Sampling (DSS) to select hard negatives during training, thereby improving optimization efficiency. However, certain datasets, such as University-1652~\citep{zheng2020university}, do not provide GPS metadata. Therefore, GLEAM-C relies solely on DSS. In our experiments, we evaluate both convolutional and transformer-based architectures as backbones. Specifically, we adopt ConvNeXt~\citep{liu2022convnet}, comprising 88.6M parameters, and the Perception Encoder~\citep{bolya2025PerceptionEncoder}, comprising 0.32B parameters.

\subsection{Training Recipe}\label{sec:c-train}

We propose a novel two-phase training scheme to integrate knowledge across multiple modalities and views (Sec.~\ref{sec:two-phase}). Given the substantial data requirements of foundational model training and the limited efficiency of the original Sample4Geo codebase, we implement targeted optimizations to significantly enhance training efficiency (Sec.~\ref{sec:dp-ddp}).

\subsubsection{Two-Phase Training Strategy}\label{sec:two-phase}

We note that a naive concatenation of multiple datasets for training poses significant challenges. Due to the backbone model’s limited matching capacity and the varying dataset sizes and difficulty levels, direct multi-dataset training often results in highly imbalanced performance. To mitigate this issue, we first novelly train the model on a single dataset that is relatively large and of moderate difficulty, allowing the backbone to acquire fundamental CVGL capabilities before introducing more complex and diverse data. In our design, we first train the model on the VIGOR~\citep{zhu2021vigor} dataset for 40 epochs, followed by training on the concatenated datasets for an additional 40 epochs. 

\subsubsection{Training Efficiency Improvement}\label{sec:dp-ddp}

Similar to CLIP training, contrastive learning models typically require large batch sizes. The original Sample4Geo adopts PyTorch’s Data Parallel (DP) strategy, where GPU 0 distributes data, aggregates outputs, and performs gradient updates for the entire cluster. This centralized design creates a significant communication bottleneck on GPU 0, thus slowing the training process. Distributed Data Parallel (DDP) alleviates this issue by allowing each GPU to compute gradients locally and synchronizing them via an efficient all-reduce. However, contrastive learning relies on in-batch negatives, and small per-GPU batch sizes limit the number of negatives available per device, potentially degrading performance. This underscores the importance of maintaining sufficiently large per-GPU batch sizes even when using DDP. To tackle this issue, we adopt the strategy implemented in OpenCLIP~\citep{ilharco_gabriel_2021_5143773}. Before computing the contrastive loss, each GPU (rank) gathers the feature representations from all other GPUs. These features are then re-concatenated, effectively creating a unified global batch for the loss calculation on each device. This process preserves the gradient flow for proper backpropagation while providing every GPU with a much larger and more effective set of negative samples. We provide pseudocode for the loss computation in Alg.~\ref{alg:loss} in Sec.~\ref{sec:ddo-loss}. This implementation achieves over 5-fold faster training speed while maintaining the original model accuracy.

\section{GLEAM-X: A Multi-Image MLLM Explanation Benchmark}~\label{sec:gleam-x}

{\vspace*{-1.2em}}

GLEAM-X extends GLEAM-C to a multi-image reasoning setting, where the model not only predicts matches between query and reference images but also generates interpretable explanations. In this section, we present the data composition (Sec.~\ref{sec:x-comp}), annotation procedure (Sec.~\ref{sec:x-anno}), evaluation protocol (Sec.~\ref{sec:x-eval}){, and how it can be combined with GLEAM-C as an integrated pipeline (Sec.~\ref{sec:integrate_pipe}).}

\subsection{Data Composition}\label{sec:x-comp}

Our training and testing data pairs are derived from the GLEAM-C training and test sets. To maintain balance across modalities and views, we perform data sampling to ensure that the amount of training data for each modality is roughly equal, as shown in Tab.~\ref{tab:sample-ratio}. In the evaluation of GLEAM-X, we also maintain this balance across modalities. The selected query images for testing include 128 for MAP, 126 for SetVL-480K, 127 for University-1652, and 123 for VIGOR. For each query image, we select one positive and one negative reference image. Using the corresponding labels, we generate explanation annotations in both English and Chinese with commercial MLLMs, specifically GPT-4o~\citep{hurst2024gpt} and Doubao-1.5-Thinking-Vision-Pro~\citep{doubao2025}. {Notably, to ensure linguistic consistency and semantic equivalence in the bilingual benchmark, we generate both language versions simultaneously in a single prompt, explicitly requiring the model to provide corresponding explanations in English and Chinese.}

During explanation construction, commercial MLLMs are provided with ground-truth labels indicating whether a given query-reference pair matches, guiding them to generate explanations accordingly. This introduces explicit prior knowledge, thereby enhancing the correctness of the generated responses. During training, the ground-truth label is concatenated with the explanations produced by the commercial MLLMs and used as the target output. For example, for a positive sample pair, the MLLM is trained to produce the response \verb|[[1]]\n\n(explanation)|. Such a format enables straightforward extraction of match labels, supporting both training supervision and evaluation of matching performance. Using this approach, we construct 200k training pairs with GPT-4o and Doubao-1.5-Thinking-Vision-Pro, respectively, and 2k test pairs with GPT-4o. The architecture for MLLM training and inference is shown in Fig.~\ref{fig:method}B.

\subsection{Data Annotation}\label{sec:x-anno}

To ensure the accuracy of the test set, we manually revise the 2k test pairs. We engage 4 human experts in Remote Sensing, each holding at least a master’s degree, to perform data annotation and revision. We conduct two annotation rounds: in each round, we first verify the correctness of the explanations generated by the commercial MLLMs and correct them if necessary; if the four experts cannot determine whether a query and reference image match, we replace the pair with a suitable one from the remaining test data and perform model-assisted generation followed by manual correction. This ensures that all test pairs are valid and unambiguous, providing a reliable basis for evaluation. More details of human annotation are provided in Sec.~\ref{sec:human_detail}.

\begin{table*}[t]\tiny
\centering
\caption{Image retrieval accuracy across backbones and training strategies. For University-1652, the last evaluation metric is AP; for all other datasets, it is Hit Rate. The first three rows (highlighted in gray) are directly taken from the corresponding papers with GPS (if applicable, marked with $^\dagger$) and DSS sampling for reference. Using a two-phase training strategy yields relatively stronger performance. The unified GLEAM-C model achieves results comparable to or even exceeding those obtained by single-dataset training.}
\renewcommand\arraystretch{0.5} %
\resizebox{0.9\textwidth}{!}{%
\centering
\setlength{\tabcolsep}{4pt} %
\begin{tabular}[\textwidth]{@{}lllcccccc@{}}
\toprule
\textbf{Backbone} & \textbf{Training Strategy} & \textbf{Dataset} & \textbf{Epochs} & \textbf{R@1} & \textbf{R@5} & \textbf{R@10} & \textbf{Top-1} & \textbf{AP/HR} \\
\midrule

\rowcolor{gray!10}
& Single-dataset & $^\dagger$VIGOR & 40 & 77.86 &  95.66 &  97.21 & 99.61 & 89.82 \\ 
\rowcolor{gray!10}
\textbf{ConvNeXt-B-384} & \textbf{Sample4Geo} & University-1652 & 1 & 92.65 & - & - & - & 93.81 \\ 
\rowcolor{gray!10}
& \textbf{$^\dagger$GPS + DSS} & $^\dagger$SetVL-480K & 40 & 16.86 &  38.95 & 47.71 & 85.40 & 16.86 \\ 
\midrule

\multirow{12}{*}{\textbf{ConvNeXt-B-384}} 
& \multirow{4}{*}{\makecell[l]{Single-dataset\\\textbf{Sample4Geo}\\\textbf{w/o GPS}}}
& VIGOR & 40 & 76.60 & 95.07 & 96.83 & 99.64 & 88.29 \\
& & University-1652 & 1 & 91.55 & 97.79 & 98.37 & 98.44 & 92.98 \\
& & MAP & 40 & 92.60 & 98.08 & 98.94 & 99.80 & 92.60 \\
& & SetVL-480K & 40 & 14.23 & 34.10 & 44.63 & 71.67 & 14.23 \\
\cmidrule(lr){2-9}

& \multirow{4}{*}{\makecell[l]{From-scratch\\(Merge)}} 
& VIGOR & \multirow{4}{*}{40} & 73.37 & 93.43 & 95.66 & 99.61 & 84.43 \\
& & University-1652 &  & 87.08 & 95.22 & 96.59 & 96.79 & 88.96 \\
& & MAP &  & 92.52 & 97.81 & 98.79 & 99.80 & 92.52 \\
& & SetVL-480K &  & 14.33 & 34.30 & 44.77 & 70.75 & 14.33 \\
\cmidrule(lr){2-9}

& \multirow{4}{*}{\makecell[l]{Two-phase\\(VIGOR $\rightarrow$ Merge)}} 
& VIGOR & \multirow{4}{*}{40} & 75.66 & 94.51 & 96.43 & 99.65 & 86.80 \\
& & University-1652 &  & 87.03 & 96.11 & 97.84 & 98.03 & 89.12 \\
& & MAP &  & 94.05 & 98.04 & 98.71 & 99.65 & 94.05 \\
& & SetVL-480K &  & 15.28 & 35.49 & 45.71 & 71.30 & 15.28 \\
\midrule

\multirow{12}{*}{\textbf{PE-Core-L14-336}} 
& \multirow{4}{*}{Single-dataset} 
& VIGOR & 40 & 75.53 & 95.48 & 97.18 & 99.67 & 89.46 \\
& & University-1652 & 1 & 94.38 & 98.44 & 98.80 & 98.85 & 95.32 \\
& & MAP & 40 & 92.79 & 97.92 & 98.55 & 99.73 & 92.79 \\
& & SetVL-480K & 40 & 21.34 & 46.54 & 57.95 & 81.58 & 21.34 \\
\cmidrule(lr){2-9}

& \multirow{4}{*}{\makecell[l]{From-scratch\\(Merge)}} 
& VIGOR & \multirow{4}{*}{40} & 69.44 & 92.81 & 95.43 & 99.63 & 83.18 \\
& & University-1652 &  & 93.20 & 97.73 & 98.30 & 98.40 & 94.26 \\
& & MAP &  & 93.11 & 98.32 & 98.75 & 99.73 & 93.11 \\
& & SetVL-480K &  & 22.00 & 46.93 & 58.28 & 81.83 & 22.00 \\
\cmidrule(lr){2-9}

& \multirow{4}{*}{\makecell[l]{Two-phase\\(VIGOR $\rightarrow$ Merge)}} 
& VIGOR & \multirow{4}{*}{40} & 75.96 & 95.46 & 97.17 & 99.68 & 89.44 \\
& & University-1652 &  & 93.19 & 97.92 & 98.49 & 98.57 & 94.28 \\
& & MAP &  & 93.97 & 98.20 & 98.43 & 99.77 & 93.97 \\
& & SetVL-480K &  & 23.25 & 48.61 & 59.76 & 82.98 & 23.25 \\
\bottomrule
\end{tabular}
}
\label{tab:all-c-results}
\vspace{-0.5em}
\end{table*}

\begin{table*}[t]\tiny
\caption{Image retrieval accuracy comparison between DP and DDP on the VIGOR dataset with ConvNeXt-B backbone using the Sample4Geo codebase. Both approaches achieve comparable accuracy, while DDP attains over 5$\times$ faster training speed.}
\vspace{-1em}
\centering
\renewcommand\arraystretch{0.1} %
\resizebox{0.9\textwidth}{!}{%
\begin{tabular}{lccccccc}
\midrule
\textbf{Dataset} & \textbf{Mode} & \textbf{Recall@1} & \textbf{Recall@5} & \textbf{Recall@10} & \textbf{Recall@top1} & \textbf{Hit Rate} & \textbf{Seconds/Epoch} \\ \midrule
\multirow{2}{*}{\textbf{VIGOR}} & {DP} & 76.80 & 95.06 & 96.70 & 99.25 & 89.05 & 521 \\ \cmidrule{2-8} 
 & {DDP} & 76.60 & 95.07 & 96.83 & 99.64 & 88.29 & 92 \\ \midrule
\end{tabular}%
}
\vspace{-2em}
\label{tab:c-speed}
\end{table*}

\subsection{Evaluation Protocol}\label{sec:x-eval}

We evaluate the MLLM outputs on the test set from two perspectives. First, we extract the predicted match labels using regular-expression parsing and compute the matching accuracy. Second, we assess the quality of the generated explanations by measuring their semantic similarity to the annotated references with Sentence-BERT~\citep{reimers2019sentence}. This dual evaluation framework ensures a comprehensive assessment, as it captures both the correctness of structured match predictions and the semantic fidelity of explanatory outputs. 

\subsection{{GLEAM-C and GLEAM-X as a Pipeline}}\label{sec:integrate_pipe}

{In real-world deployment, GLEAM-C and GLEAM-X operate as a two-stage pipeline that combines efficient image retrieval with interpretable verification, as illustrated in Fig.~\ref{fig:overview}. Given a query image and a reference database, GLEAM-C first retrieves the most similar reference image through CVGL. GLEAM-X then verifies whether the query and retrieved reference correspond to the same geographic location, while providing human-interpretable explanations that identify key visual correspondences or discrepancies between the image pair. This integrated approach not only enhances retrieval reliability but also offers transparency through explainable reasoning, which is critical for safety-critical real-world applications such as autonomous navigation, search and rescue operations, and emergency response.}

\section{Experiments}\label{sec:exp}

In this section, we design a series of experiments to evaluate the unified GLEAM-C model (Sec.~\ref{sec:res-c}) and the GLEAM-X benchmark (Sec.~\ref{sec:res-x}). All training experiments are conducted on a cluster with 10 NVIDIA RTX 4090 D GPUs, each with 48 GB of memory. In addition, to assess practical deployment feasibility, we evaluate the models' inference performance on an NVIDIA Jetson AGX Xavier for real-world edge computing scenarios in the Appendix (Sec.~\ref{sec:deployment}).

\subsection{Evaluation Results of GLEAM-C}\label{sec:res-c}

For GLEAM-C, we compare the unified model with single-dataset training and evaluate the data integration strategy (Sec.\ref{sec:c-inter}). We further assess improvements in training efficiency (Sec.\ref{sec:c-efficiency}) relative to prior SOTA training methods. Additionally, we investigate cross-domain generalization performance and the rationale for selecting VIGOR as the initial training dataset (Sec.~\ref{sec:dis_gleamc}).

\begin{table*}[t]\tiny
\centering
\caption{Matching accuracy comparison of different models on GLEAM-X in English and Chinese. Results are reported on MAP, SetVL-480K, University-1652, and VIGOR datasets, along with the overall average accuracy. Label-only supervision and LLM-explanation supervision (+GPT-4o, +Doubao-1.5) significantly improve the performance of Qwen2.5-VL-3B-Instruct.}
\renewcommand\arraystretch{0.5} %
\vspace{-1em}
\resizebox{0.9\textwidth}{!}{%
\begin{tabular}{lcccccc}
\midrule
\textbf{Model} & \textbf{Language} & \textbf{MAP} & \textbf{SetVL-480K} & \textbf{University-1652} & \textbf{VIGOR} & \textbf{Avg Acc} \\ \midrule
{GPT-4o} &  & 85.55 & 63.89 & 93.70 & 81.71 & 81.25 \\ %
{Doubao-1.5} &  & 91.80 & 74.21 & 60.24 & 83.74 & 77.48 \\ \cmidrule{1-1} \cmidrule{3-7} 
{Qwen2.5-VL-3B-Instruct} &  & 55.08 & 57.94 & 51.57 & 65.04 & 57.34 \\ %
{\quad+label only} &  & 96.88 & 84.13 & 97.64 & 92.28 & 92.76 \\ %
{\quad+GPT-4o} &  & 94.53 & 78.17 & 99.61 & 89.02 & 90.38 \\ %
{\quad+Doubao-1.5} & \multirow{-6}{*}{EN} & 95.31 & 79.37 & 96.46 & 87.80 & 89.78 \\ \midrule
\textbf{Model} & \textbf{Language} & \textbf{MAP} & \textbf{SetVL-480K} & \textbf{University-1652} & \textbf{VIGOR} & \textbf{Avg Acc} \\ \midrule
{GPT-4o} &  & 85.55 & 70.63 & 96.06 & 82.11 & 83.63 \\ %
{Doubao-1.5} &  & 89.45 & 78.97 & 66.93 & 81.30 & 79.17 \\ \cmidrule{1-1} \cmidrule{3-7} 
{Qwen2.5-VL-3B-Instruct} &  & 51.17 & 51.19 & 50.00 & 49.59 & 50.50 \\ %
{\quad+label only} &  & 96.88 & 84.13 & 97.64 & 92.28 & 92.76 \\ %
{\quad+GPT-4o} &  & 95.31 & 76.59 & 99.61 & 89.02 & 90.18 \\ %
{\quad+Doubao-1.5} & \multirow{-6}{*}{ZH} & 94.53 & 75.00 & 94.88 & 86.59 & 87.80 \\ \midrule
\end{tabular}%
}
\vspace{-2em}
\label{tab:x-acc}
\end{table*}

\subsubsection{From Single Dataset to Multi-Dataset}\label{sec:c-inter}

As in most CVGL studies, we first train on single-view and single-modality datasets. Training is conducted separately on the VIGOR (same), University-1652, MAP, and SetVL-480K (N = 1) datasets. Model performance is evaluated using image retrieval accuracy, with University-1652 following the Drone2Sat protocol. We adopt ConvNeXt-B-384 (88.6M) and PE-Core-L14-336 (0.32B) as the image encoders. The results for single-dataset training are presented in Tab.~\ref{tab:all-c-results}. Since some datasets lack GPS metadata, we do not use GPS sampling in all our model training, which may result in a slight decrease in accuracy. For reference, we also report the original metrics from the respective papers, with GPS (if applicable) and DSS sampling~\citep{deuser2023sample4geo,Wu2024CrossViewIS}.

To train the unified model, we compare two data integration strategies in our work. The first strategy trains on mixed data directly from the backbone’s initial weights, while the second strategy starts training from a VIGOR checkpoint obtained through single-dataset training. We use the data sampling ratios shown in Tab.~\ref{tab:sample-ratio}, and both strategies are trained for 40 epochs. We provide the results in Tab.~\ref{tab:all-c-results} and share the following observations:

\textbf{1)} Directly training on the mixed data degrades the performance on VIGOR. However, continuing training the unified model from a VIGOR checkpoint yields better results and improves accuracy on MAP and SetVL-480K. This highlights the importance of endowing the model with basic CVGL capability before performing mixed-data training, and also suggests that learning across different views and modalities may mutually reinforce each other.

\textbf{2)} Compared with single-dataset training, our novel two-phase training strategy yields the foundational GLEAM-C model, which attains comparable or superior performance across all evaluated datasets.

\textbf{3)} The larger PE-Core-L14-336 model achieves overall better performance compared to ConvNeXt-B-384 after training. Additionally, although University-1652 requires only a single epoch of training, we train for 40 epochs due to the demands of mixed-data training; under this setting, smaller models may suffer from overtraining or instability, whereas larger models remain largely unaffected.

\subsubsection{Training Efficiency Evaluation}\label{sec:c-efficiency}

We evaluate training efficiency on the VIGOR dataset by comparing DP and DDP strategies. Following the experiment settings of Sample4Geo~\citep{deuser2023sample4geo}, we adopt the ConvNeXt-B-384 backbone and use the original Sample4Geo code for DP training (without GPS-based sampling). As shown in Tab.~\ref{tab:c-speed}, both approaches achieve comparable image retrieval accuracy across all Recall@k and Hit Rate metrics. However, DDP demonstrates a substantial (5-fold) speed advantage, reducing the time per training epoch from 521 seconds to 92 seconds, highlighting its effectiveness for accelerating large-scale CVGL model training. 

\subsubsection{{Discussion: Cross-Domain Generalization \& Initial Dataset Selection}~\label{sec:dis_gleamc}
}

{Beyond the above analysis, we further investigate two questions regarding our unified model training scheme: \textbf{1)} What is the cross-domain generalization performance when trained on one modality and tested on unseen modalities? \textbf{2)} Why VIGOR is chosen as the initial dataset in our two-phase training strategy? Detailed experimental analysis and results are provided in Sec.~\ref{app:cross_domain} and Sec.~\ref{app:vigor_selection}, respectively.}

\subsection{Evaluation Results on GLEAM-X}\label{sec:res-x}

We evaluate both commercial models and open-source models on GLEAM-X. For commercial models, we regenerate answers on the test set using GPT-4o and Doubao-1.5-Thinking-Vision-Pro without providing ground-truth labels. For open-source models, we compare three variants of Qwen2.5-VL-3B-Instruct: the original model, the version fine-tuned with LLM-explanation supervision, and a version trained solely with label-only supervision. Specifically, for the explanation-based variant, we follow the procedure in Sec.~\ref{sec:x-comp}, concatenating labels and explanations during training. We train separately on the 200k explanations generated by GPT-4o and Doubao-1.5 for 1 epoch. Conversely, the label-only variant uses only labels (e.g., [[1]]) for training.

After inference on the test set, we decouple matching predictions and explanations through regularization to isolate their respective contributions. We provide a comprehensive evaluation of GLEAM-X through detailed analyses on matching accuracy (Sec.\ref{sec:x-acc}) and semantic accuracy (Sec.\ref{sec:x-sim}). We further investigate positive/negative sample characteristics (Sec.~\ref{sec:x-neg}), explore LLM-as-a-judge and human test beyond Sentence-BERT (Sec.~\ref{sec:main_judge}), and discuss potential biases in data curation (Sec.~\ref{sec:main_dis_x}).

\begin{table*}[t]\tiny
\centering
\renewcommand\arraystretch{0.5} %
\caption{Similarity score distribution (by Sentence-BERT) for different models on GLEAM-X in English and Chinese. Columns show percentages in each similarity range (0.0–0.2, ..., 0.8–1.0) and the last column reports Avg Sim. LLM-explanation supervision (+GPT-4o, +Doubao-1.5) improves alignment with ground-truth, while label-only supervision fails to generate explanations.}
\vspace{-0.5em}
\resizebox{0.9\textwidth}{!}{%
\begin{tabular}{lc|ccccc|c}
\midrule
{\textbf{Model}} & {\textbf{Language}} & {\textbf{0.0-0.2}} & {\textbf{0.2-0.4}} & {\textbf{0.4-0.6}} & {\textbf{0.6-0.8}} & {\textbf{0.8-1.0}} & {\textbf{Avg Sim}} \\ \midrule
{GPT-4o} &  & 0.0\% & 0.1\% & 0.8\% & 21.6\% & 77.5\% & 0.8349 \\ %
{Doubao-1.5} &  & 0.0\% & 0.0\% & 0.2\% & 33.5\% & 66.3\% & 0.8171 \\ \cmidrule{1-1} \cmidrule{3-8}
{Qwen2.5-VL-3B-Instruct} &  & 0.8\% & 0.0\% & 2.0\% & 59.3\% & 37.9\% & 0.7645 \\ %
{\quad+label only} &  & 99.9\% & 0.1\% & 0.0\% & 0.0\% & 0.0\% & 0.0981 \\ %
{\quad+GPT-4o} &  & 0.0\% & 0.0\% & 0.1\% & 22.4\% & 77.5\% & 0.8405 \\ %
{\quad+Doubao-1.5} & \multirow{-6}{*}{EN} & 0.0\% & 0.0\% & 0.5\% & 36.0\% & 63.5\% & 0.8122 \\ \midrule
{\textbf{Model}} & {\textbf{Language}} & {\textbf{0.0-0.2}} & {\textbf{0.2-0.4}} & {\textbf{0.4-0.6}} & {\textbf{0.6-0.8}} & {\textbf{0.8-1.0}} & {\textbf{Avg Sim}} \\ \midrule
{GPT-4o} &  & 1.6\% & 5.0\% & 2.3\% & 50.5\% & 40.7\% & 0.7403 \\ %
{Doubao-1.5} &  & 0.0\% & 0.0\% & 4.5\% & 66.7\% & 28.9\% & 0.7495 \\ \cmidrule{1-1} \cmidrule{3-8} 
{Qwen2.5-VL-3B-Instruct} &  & 4.2\% & 0.1\% & 2.9\% & 58.5\% & 34.3\% & 0.7399 \\ %
{\quad+label only} &  & 92.1\% & 7.9\% & 0.0\% & 0.0\% & 0.0\% & 0.1476 \\ %
{\quad+GPT-4o} &  & 0.0\% & 0.0\% & 0.3\% & 32.5\% & 67.2\% & 0.8239 \\ %
{\quad+Doubao-1.5} & \multirow{-6}{*}{ZH} & 0.4\% & 0.0\% & 1.1\% & 46.6\% & 51.9\% & 0.7924 \\ \midrule
\end{tabular}%
}
\vspace{-1em}
\label{tab:x-sim}
\end{table*}

\begin{table}[h]\small
\centering
\caption{Positive and negative sample accuracy on GLEAM-X. The Pos-Neg columns report the difference between positive and negative samples. Vanilla Qwen2.5-VL-3B-Instruct shows a strong bias toward predicting non-matching pairs, while model fine-tuning (+GPT-4o, +Doubao-1.5) mitigates this bias.}
\renewcommand\arraystretch{0.8} %
\resizebox{\linewidth}{!}{%
\begin{tabular}{lcccc}
\midrule
{\textbf{Model}} & {\textbf{Lang}} & {\textbf{Pos Acc}} & {\textbf{Neg Acc}} & {\textbf{Pos-Neg}}  \\ \midrule
{GPT-4o} &  & 70.83 & 91.67 & -20.84 \\ %
{Doubao-1.5} &  & 66.87 & 88.10 & -21.23  \\ \cmidrule{1-1} \cmidrule{3-5} 
{Qwen2.5-VL-3B} &  & 23.81 & 90.87 & -67.06 \\ %
{\quad+label only} &  & 93.85 & 91.67 & 2.18  \\ %
{\quad+GPT-4o} &  & 93.25 & 87.50 & 5.75  \\ %
{\quad+Doubao-1.5} & \multirow{-6}{*}{EN} & 90.08 & 89.48 & 0.60 \\ \midrule
{\textbf{Model}} & {\textbf{Lang}} & {\textbf{Pos Acc}} & {\textbf{Neg Acc}} & {\textbf{Pos-Neg}}  \\ \midrule
{GPT-4o} &  & 73.02 & 94.25 & -21.23  \\ %
{Doubao-1.5} &  & 76.19 & 82.14 & -5.95  \\ \cmidrule{1-1} \cmidrule{3-5} 
{Qwen2.5-VL-3B} &  & 1.19 & 99.80 & -98.61 \\ %
{\quad+label only} &  & 93.85 & 91.67 & 2.18 \\ %
{\quad+GPT-4o} &  & 90.87 & 89.48 & 1.39  \\ %
{\quad+Doubao-1.5} & \multirow{-6}{*}{ZH} & 84.52 & 91.07 & -6.55 \\ \midrule
\end{tabular}%
}
\vspace{-2em}
\label{tab:x-neg}
\end{table}

\subsubsection{Matching Accuracy}\label{sec:x-acc}

Tab.~\ref{tab:x-acc} presents the matching accuracy results. Commercial models achieve approximately 80\% accuracy, with GPT-4o outperforming Doubao-1.5 on average. In contrast, the original Qwen2.5-VL-3B-Instruct model exhibits significantly lower accuracy, particularly in Chinese (50.50\%). After fine-tuning on the explanation dataset, all model variants surpass commercial MLLMs. Notably, the label-only supervision model achieves the highest accuracy, likely because its training targets are extremely short (e.g., [[0]] or [[1]]), reducing learning difficulty. {By comparison, the explanation-supervised model must learn the harder task of generating full reasoning while making predictions, resulting in relatively lower accuracy. Despite this, it still substantially outperforms the baseline while providing interpretable explanations for its decisions. This reflects a design choice where GLEAM-X prioritizes interpretable geo-localization over maximizing matching accuracy alone.}

\subsubsection{Semantic Analysis of Explanations}\label{sec:x-sim}

Tab.~\ref{tab:x-sim} reveals the similarity distributional differences across models. Commercial MLLMs concentrate in the high-similarity range (0.8–1.0) for English, but shift toward 0.6–0.8 in Chinese. The original Qwen2.5-VL-3B places a substantial proportion of predictions in the mid-similarity bins (0.6–0.8) for both languages, indicating weaker semantic alignment with ground-truth explanations. Fine-tuning with LLM-explanation supervision shifts the distribution upward, with both +GPT-4o and +Doubao-1.5 variants peaking in the 0.8–1.0 range. By contrast, the label-only supervision model collapses almost entirely into the lowest bin (0.0–0.2), confirming its inability to generate semantically meaningful explanations despite improvements in matching accuracy.

\subsubsection{Positive and Negative Pair Samples}\label{sec:x-neg}

To further analyze model behavior on positive and negative samples, we report matching accuracy in Tab.~\ref{tab:x-neg}. Vanilla MLLMs, including both commercial models and Qwen2.5-VL-3B-Instruct, exhibit a bias toward negative predictions. Notably, Qwen2.5-VL-3B attains very low accuracy on positive samples while maintaining high accuracy on negative samples, suggesting that the original model tends to classify image pairs as non-matching. This negative bias likely arises from an inherent asymmetry of the task: identifying mismatches can be triggered by a single discrepant cue, whereas confirming a match requires consistent evidence across multiple aspects. After fine-tuning, this bias is substantially alleviated. Positive accuracy increases to over 90\% in English and over 84\% in Chinese, while negative accuracy remains high. These results indicate that fine-tuning effectively improves the model’s ability to verify matches without sacrificing its capability to detect mismatches.

\subsubsection{{Discussion: More Semantic Analysis}\label{sec:main_judge}}
{
For semantic analysis, while Sentence-BERT offers low validation costs in terms of both price and computational efficiency, it is limited to measuring textual similarity between model outputs and annotations, potentially failing to capture critical reasoning steps grounded in the visual content itself. To address this limitation, we supplement our evaluation with an LLM-as-a-judge approach using the multimodal Gemini 2.5 Flash~\citep{comanici2025gemini} model, which can assess both textual reasoning and visual matching. Furthermore, we conduct a human test to validate the effectiveness of both automated assessment methods. Our experiments demonstrate strong alignment across all three evaluation approaches (detailed in Sec.~\ref{sec:more_judge}).
}

\section{Conclusion}\label{sec:conclusion}

In this paper, we systematically address the challenges of CVGL through several key contributions. First, we provide a detailed analysis of the CVGL problem, highlighting the benefits of a unified framework for integrating multiple views and modalities, and emphasizing the importance of interpretability in practical applications. Second, we propose GLEAM-C, a novel foundational model that aligns diverse viewpoints and modalities with satellite imagery through a two-phase training strategy, achieving high accuracy and training efficiency. Third, we introduce GLEAM-X, a novel bilingual benchmark enabling explainable reasoning with human-interpretable explanations for image correspondences. GLEAM-C and GLEAM-X further integrate into a unified pipeline combining accurate correspondence prediction with interpretable explanations, improving both robustness and trustworthiness of CVGL systems.

\section*{Impact Statements}

This paper advances cross-view geo-localization (CVGL) and explainable AI, with applications in autonomous navigation, robotics, and disaster response. The core CVGL model establishes robust cross-modal alignment within a unified framework, enabling efficient deployment on resource-constrained platforms such as drones and vehicles. The explainability component enhances transparency in safety-critical systems. As with many technologies in computer vision and geo-localization, there exist potential considerations regarding responsible use, including concerns related to geographic privacy information. All datasets used in this work are publicly available and widely adopted in the research community. In the spirit of open science and to facilitate reproducible research, we commit to releasing all code, trained models, and datasets publicly. We encourage responsible deployment with appropriate consideration of ethical guidelines.

\bibliography{example_paper}

@article{chen2025multi,
  title={Multi-level embedding and alignment network with consistency and invariance learning for cross-view geo-localization},
  author={Chen, Zhongwei and Yang, Zhao-Xu and Rong, Hai-Jun},
  journal={IEEE Transactions on Geoscience and Remote Sensing},
  year={2025},
  publisher={IEEE}
}

@article{wang2024multiple,
  title={Multiple-environment self-adaptive network for aerial-view geo-localization},
  author={Wang, Tingyu and Zheng, Zhedong and Sun, Yaoqi and Yan, Chenggang and Yang, Yi and Chua, Tat-Seng},
  journal={Pattern Recognition},
  volume={152},
  pages={110363},
  year={2024},
  publisher={Elsevier}
}

@article{yao2024minicpm,
  title={Minicpm-v: A gpt-4v level mllm on your phone},
  author={Yao, Yuan and Yu, Tianyu and Zhang, Ao and Wang, Chongyi and Cui, Junbo and Zhu, Hongji and Cai, Tianchi and Li, Haoyu and Zhao, Weilin and He, Zhihui and others},
  journal={arXiv preprint arXiv:2408.01800},
  year={2024}
}

@article{hurst2024gpt,
  title={Gpt-4o system card},
  author={Hurst, Aaron and Lerer, Adam and Goucher, Adam P and Perelman, Adam and Ramesh, Aditya and Clark, Aidan and Ostrow, AJ and Welihinda, Akila and Hayes, Alan and Radford, Alec and others},
  journal={arXiv preprint arXiv:2410.21276},
  year={2024}
}

@inproceedings{Achiam2023GPT4TR,
  title={GPT-4 Technical Report},
  author={OpenAI},
  journal={ArXiv},
  year={2023},
  volume={abs/2303.08774}
}

@inproceedings{chen2024internvl,
  title={Internvl: Scaling up vision foundation models and aligning for generic visual-linguistic tasks},
  author={Chen, Zhe and Wu, Jiannan and Wang, Wenhai and Su, Weijie and Chen, Guo and Xing, Sen and Zhong, Muyan and Zhang, Qinglong and Zhu, Xizhou and Lu, Lewei and others},
  booktitle={Proceedings of the IEEE/CVF Conference on Computer Vision and Pattern Recognition},
  pages={24185--24198},
  year={2024}
}

@article{Bai2025Qwen25VLTR,
  title={Qwen2.5-VL Technical Report},
  author={Shuai Bai and Keqin Chen and Xuejing Liu and Jialin Wang and Wenbin Ge and Sibo Song and Kai Dang and Peng Wang and Shijie Wang and Jun Tang and Humen Zhong and Yuanzhi Zhu and Mingkun Yang and Zhaohai Li and Jianqiang Wan and Pengfei Wang and Wei Ding and Zheren Fu and Yiheng Xu and Jiabo Ye and Xi Zhang and Tianbao Xie and Zesen Cheng and Hang Zhang and Zhibo Yang and Haiyang Xu and Junyang Lin},
  journal={ArXiv},
  year={2025},
  volume={abs/2502.13923},
  url={https://api.semanticscholar.org/CorpusID:276449796}
}

@article{li2024llava,
  title={Llava-onevision: Easy visual task transfer},
  author={Li, Bo and Zhang, Yuanhan and Guo, Dong and Zhang, Renrui and Li, Feng and Zhang, Hao and Zhang, Kaichen and Li, Yanwei and Liu, Ziwei and Li, Chunyuan},
  journal={arXiv preprint arXiv:2408.03326},
  year={2024}
}

@article{achiam2023gpt4,
  title={{GPT-4} technical report},
  author={Achiam, Josh and Adler, Steven and Agarwal, Sandhini and Ahmad, Lama and Akkaya, Ilge and Aleman, Florencia Leoni and Almeida, Diogo and Altenschmidt, Janko and Altman, Sam and Anadkat, Shyamal and others},
  journal={arXiv preprint arXiv:2303.08774},
  year={2023}
}

@article{Zhu2025InternVL3EA,
  title={InternVL3: Exploring Advanced Training and Test-Time Recipes for Open-Source Multimodal Models},
  author={Jinguo Zhu and Weiyun Wang and Zhe Chen and Zhaoyang Liu and Shenglong Ye and Lixin Gu and Yuchen Duan and Hao Tian and Weijie Su and Jie Shao and Zhangwei Gao and Erfei Cui and Yue Cao and Yangzhou Liu and Haomin Wang and Weiye Xu and Hao Li and Jiahao Wang and Han Lv and Dengnian Chen and Songze Li and Yinan He and Tan Jiang and Jiapeng Luo and Yi Wang and Cong He and Botian Shi and Xingcheng Zhang and Wenqi Shao and Junjun He and Ying Xiong and Wenwen Qu and Peng Sun and Penglong Jiao and Lijun Wu and Kai Zhang and Hui Deng and Jiaye Ge and Kaiming Chen and Limin Wang and Min Dou and Lewei Lu and Xizhou Zhu and Tong Lu and Dahua Lin and Yu Qiao and Jifeng Dai and Wenhai Wang},
  journal={ArXiv},
  year={2025},
  volume={abs/2504.10479},
  url={https://api.semanticscholar.org/CorpusID:277780955}
}

@article{wang2025internvl3,
  title={InternVL3. 5: Advancing Open-Source Multimodal Models in Versatility, Reasoning, and Efficiency},
  author={Wang, Weiyun and Gao, Zhangwei and Gu, Lixin and Pu, Hengjun and Cui, Long and Wei, Xingguang and Liu, Zhaoyang and Jing, Linglin and Ye, Shenglong and Shao, Jie and others},
  journal={arXiv preprint arXiv:2508.18265},
  year={2025}
}

@article{chen2025janus,
  title={Janus-Pro: Unified Multimodal Understanding and Generation with Data and Model Scaling},
  author={Chen, Xiaokang and Wu, Zhiyu and Liu, Xingchao and Pan, Zizheng and Liu, Wen and Xie, Zhenda and Yu, Xingkai and Ruan, Chong},
  journal={arXiv preprint arXiv:2501.17811},
  year={2025}
}

@inproceedings{workman2015wide,
  title={Wide-area image geolocalization with aerial reference imagery},
  author={Workman, Scott and Souvenir, Richard and Jacobs, Nathan},
  booktitle={Proceedings of the IEEE International Conference on Computer Vision},
  pages={3961--3969},
  year={2015}
}

@inproceedings{liu2019lending,
  title={Lending orientation to neural networks for cross-view geo-localization},
  author={Liu, Liu and Li, Hongdong},
  booktitle={Proceedings of the IEEE/CVF conference on computer vision and pattern recognition},
  pages={5624--5633},
  year={2019}
}

@article{deng2025bagel,
  title   = {Emerging Properties in Unified Multimodal Pretraining},
  author  = {Deng, Chaorui and Zhu, Deyao and Li, Kunchang and Gou, Chenhui and Li, Feng and Wang, Zeyu and Zhong, Shu and Yu, Weihao and Nie, Xiaonan and Song, Ziang and Shi, Guang and Fan, Haoqi},
  journal = {arXiv preprint arXiv:2505.14683},
  year    = {2025}
}

@article{xie2024show,
  title={Show-o: One single transformer to unify multimodal understanding and generation},
  author={Xie, Jinheng and Mao, Weijia and Bai, Zechen and Zhang, David Junhao and Wang, Weihao and Lin, Kevin Qinghong and Gu, Yuchao and Chen, Zhijie and Yang, Zhenheng and Shou, Mike Zheng},
  journal={arXiv preprint arXiv:2408.12528},
  year={2024}
}

@article{bai2025qwen2,
  title={Qwen2. 5-vl technical report},
  author={Bai, Shuai and Chen, Keqin and Liu, Xuejing and Wang, Jialin and Ge, Wenbin and Song, Sibo and Dang, Kai and Wang, Peng and Wang, Shijie and Tang, Jun and others},
  journal={arXiv preprint arXiv:2502.13923},
  year={2025}
}

@article{comanici2025gemini,
  title={Gemini 2.5: Pushing the frontier with advanced reasoning, multimodality, long context, and next generation agentic capabilities},
  author={Comanici, Gheorghe and Bieber, Eric and Schaekermann, Mike and Pasupat, Ice and Sachdeva, Noveen and Dhillon, Inderjit and Blistein, Marcel and Ram, Ori and Zhang, Dan and Rosen, Evan and others},
  journal={arXiv preprint arXiv:2507.06261},
  year={2025}
}

@article{liu2023llava,
  title={Visual Instruction Tuning}, 
  author={Liu, Haotian and Li, Chunyuan and Wu, Qingyang and Lee, Yong Jae},
  journal={NeurIPS},
  volume={36},
  year={2024}
}

@article{dai2023vision,
  title={Vision-based UAV self-positioning in low-altitude urban environments},
  author={Dai, Ming and Zheng, Enhui and Feng, Zhenhua and Qi, Lei and Zhuang, Jiedong and Yang, Wankou},
  journal={IEEE Transactions on Image Processing},
  volume={33},
  pages={493--508},
  year={2023},
  publisher={IEEE}
}

@inproceedings{bansal2011geo,
  title={Geo-localization of street views with aerial image databases},
  author={Bansal, Mayank and Sawhney, Harpreet S and Cheng, Hui and Daniilidis, Kostas},
  booktitle={Proceedings of the 19th ACM international conference on Multimedia},
  pages={1125--1128},
  year={2011}
}

@inproceedings{castaldo2015semantic,
  title={Semantic cross-view matching},
  author={Castaldo, Francesco and Zamir, Amir and Angst, Roland and Palmieri, Francesco and Savarese, Silvio},
  booktitle={Proceedings of the IEEE International Conference on Computer Vision Workshops},
  pages={9--17},
  year={2015}
}

@article{krizhevsky2012imagenet,
  title={Imagenet classification with deep convolutional neural networks},
  author={Krizhevsky, Alex and Sutskever, Ilya and Hinton, Geoffrey E},
  journal={Advances in neural information processing systems},
  volume={25},
  year={2012}
}

@inproceedings{shi2020optimal,
  title={Optimal feature transport for cross-view image geo-localization},
  author={Shi, Yujiao and Yu, Xin and Liu, Liu and Zhang, Tong and Li, Hongdong},
  booktitle={Proceedings of the AAAI Conference on Artificial Intelligence},
  volume={34},
  number={07},
  pages={11990--11997},
  year={2020}
}

@article{shi2019spatial,
  title={Spatial-aware feature aggregation for image based cross-view geo-localization},
  author={Shi, Yujiao and Liu, Liu and Yu, Xin and Li, Hongdong},
  journal={Advances in Neural Information Processing Systems},
  volume={32},
  year={2019}
}

@article{dai2021transformer,
  title={A transformer-based feature segmentation and region alignment method for UAV-view geo-localization},
  author={Dai, Ming and Hu, Jianhong and Zhuang, Jiedong and Zheng, Enhui},
  journal={IEEE Transactions on Circuits and Systems for Video Technology},
  volume={32},
  number={7},
  pages={4376--4389},
  year={2021},
  publisher={IEEE}
}

@article{xia2024enhancing,
  title={Enhancing cross-view geo-localization with domain alignment and scene consistency},
  author={Xia, Panwang and Wan, Yi and Zheng, Zhi and Zhang, Yongjun and Deng, Jiwei},
  journal={IEEE Transactions on Circuits and Systems for Video Technology},
  year={2024},
  publisher={IEEE}
}

@inproceedings{zheng2020university,
  title={University-1652: A multi-view multi-source benchmark for drone-based geo-localization},
  author={Zheng, Zhedong and Wei, Yunchao and Yang, Yi},
  booktitle={Proceedings of the 28th ACM international conference on Multimedia},
  pages={1395--1403},
  year={2020}
}

@inproceedings{zhu2021vigor,
  title={Vigor: Cross-view image geo-localization beyond one-to-one retrieval},
  author={Zhu, Sijie and Yang, Taojiannan and Chen, Chen},
  booktitle={Proceedings of the IEEE/CVF Conference on Computer Vision and Pattern Recognition},
  pages={3640--3649},
  year={2021}
}

@article{wu2024cross,
  title={Cross-View Image Set Geo-Localization},
  author={Wu, Qiong and Xia, Panwang and Yu, Lei and Liu, Yi and Xiong, Mingtao and Zhong, Liheng and Chen, Jingdong and Yang, Ming and Zhang, Yongjun and Wan, Yi},
  journal={arXiv preprint arXiv:2412.18852},
  year={2024}
}

@inproceedings{deuser2023sample4geo,
  title={Sample4geo: Hard negative sampling for cross-view geo-localisation},
  author={Deuser, Fabian and Habel, Konrad and Oswald, Norbert},
  booktitle={Proceedings of the IEEE/CVF International Conference on Computer Vision},
  pages={16847--16856},
  year={2023}
}

@article{shen2023mccg,
  title={MCCG: A ConvNeXt-based multiple-classifier method for cross-view geo-localization},
  author={Shen, Tianrui and Wei, Yingmei and Kang, Lai and Wan, Shanshan and Yang, Yee-Hong},
  journal={IEEE Transactions on Circuits and Systems for Video Technology},
  volume={34},
  number={3},
  pages={1456--1468},
  year={2023},
  publisher={IEEE}
}

@article{li2024learning,
  title={Learning cross-view visual geo-localization without ground truth},
  author={Li, Haoyuan and Xu, Chang and Yang, Wen and Yu, Huai and Xia, Gui-Song},
  journal={IEEE Transactions on Geoscience and Remote Sensing},
  year={2024},
  publisher={IEEE}
}

@inproceedings{lin2015learning,
  title={Learning deep representations for ground-to-aerial geolocalization},
  author={Lin, Tsung-Yi and Cui, Yin and Belongie, Serge and Hays, James},
  booktitle={Proceedings of the IEEE conference on computer vision and pattern recognition},
  pages={5007--5015},
  year={2015}
}

@article{bolya2025PerceptionEncoder,
  title={Perception Encoder: The best visual embeddings are not at the output of the network},
  author={Daniel Bolya and Po-Yao Huang and Peize Sun and Jang Hyun Cho and Andrea Madotto and Chen Wei and Tengyu Ma and Jiale Zhi and Jathushan Rajasegaran and Hanoona Rasheed and Junke Wang and Marco Monteiro and Hu Xu and Shiyu Dong and Nikhila Ravi and Daniel Li and Piotr Doll{\'a}r and Christoph Feichtenhofer},
  journal={arXiv:2504.13181},
  year={2025}
}

@misc{doubao2025,
  author = {Volcengine},
  title = {Doubao-1.5-Thinking-Vision-Pro},
  year = {2025},
  url = {https://www.volcengine.com/docs/82379/1554521},
}

@article{reimers2019sentence,
  title={Sentence-bert: Sentence embeddings using siamese bert-networks},
  author={Reimers, Nils and Gurevych, Iryna},
  journal={arXiv preprint arXiv:1908.10084},
  year={2019}
}

@software{ilharco_gabriel_2021_5143773,
  author       = {Ilharco, Gabriel and
                  Wortsman, Mitchell and
                  Wightman, Ross and
                  Gordon, Cade and
                  Carlini, Nicholas and
                  Taori, Rohan and
                  Dave, Achal and
                  Shankar, Vaishaal and
                  Namkoong, Hongseok and
                  Miller, John and
                  Hajishirzi, Hannaneh and
                  Farhadi, Ali and
                  Schmidt, Ludwig},
  title        = {OpenCLIP},
  month        = jul,
  year         = 2021,
  note         = {If you use this software, please cite it as below.},
  publisher    = {Zenodo},
  version      = {0.1},
  doi          = {10.5281/zenodo.5143773},
  url          = {https://doi.org/10.5281/zenodo.5143773}
}

@inproceedings{liu2022convnet,
  title={A convnet for the 2020s},
  author={Liu, Zhuang and Mao, Hanzi and Wu, Chao-Yuan and Feichtenhofer, Christoph and Darrell, Trevor and Xie, Saining},
  booktitle={Proceedings of the IEEE/CVF conference on computer vision and pattern recognition},
  pages={11976--11986},
  year={2022}
}

@article{Wu2024CrossViewIS,
  title={Cross-View Image Set Geo-Localization},
  author={Qiong Wu and Panwang Xia and Lei Yu and Yi Liu and Mingtao Xiong and Liheng Zhong and Jingdong Chen and Ming Yang and Yongjun Zhang and Yi Wan},
  journal={ArXiv},
  year={2024},
  volume={abs/2412.18852},
}

@inproceedings{li2024unleashing,
  title={Unleashing unlabeled data: A paradigm for cross-view geo-localization},
  author={Li, Guopeng and Qian, Ming and Xia, Gui-Song},
  booktitle={Proceedings of the IEEE/CVF Conference on Computer Vision and Pattern Recognition},
  pages={16719--16729},
  year={2024}
}

@article{cui2003autonomous,
  title={Autonomous vehicle positioning with GPS in urban canyon environments},
  author={Cui, Youjing and Ge, Shuzhi Sam},
  journal={IEEE transactions on robotics and automation},
  volume={19},
  number={1},
  pages={15--25},
  year={2003},
  publisher={IEEE}
}

@article{chen2023gnss,
  title={GNSS high-precision augmentation for autonomous vehicles: Requirements, solution, and technical challenges},
  author={Chen, Liang and Zheng, Fu and Gong, Xiaopeng and Jiang, Xinyuan},
  journal={Remote Sensing},
  volume={15},
  number={6},
  pages={1623},
  year={2023},
  publisher={MDPI}
}

@article{xu2024precise,
  title={A precise localization algorithm for unmanned aerial vehicles integrating visual-internal odometry and cartographer},
  author={Xu, Jiaqi and Chen, Zhou and Chen, Jie and Zhou, Jingyan and Du, Xiaofei},
  journal={Journal of Measurements in Engineering},
  volume={12},
  number={2},
  pages={284--297},
  year={2024},
  publisher={Extrica}
}

@inproceedings{suzuki2016precise,
  title={Precise UAV position and attitude estimation by multiple GNSS receivers for 3D mapping},
  author={Suzuki, Taro and Takahashi, Yusuke and Amano, Yoshiharu},
  booktitle={Proceedings of the 29th International Technical Meeting of the Satellite Division of The Institute of Navigation (ION GNSS+ 2016)},
  pages={1455--1464},
  year={2016}
}

@article{nowak2024enhancing,
  title={Enhancing Mobile Robot Position Estimation with Machine Learning Methods Using Camera-Based Tracking},
  author={Nowak, Tom and Gro{\ss}e-Kreul, Alexander and Boshoff, Marius and Kuhlenk{\"o}tter, Bernd},
  journal={Procedia CIRP},
  volume={130},
  pages={964--968},
  year={2024},
  publisher={Elsevier}
}

@article{semborski2024review,
  title={A review on positioning techniques of mobile robots},
  author={Semborski, Jakub and Idzkowski, Adam},
  journal={Robotic Systems and Applications},
  volume={4},
  number={1},
  pages={30--43},
  year={2024},
  publisher={Extrica}
}

@article{kamalam2022augmented,
  title={Augmented Reality-Centered Position Navigation for Wearable Devices with Machine Learning Techniques},
  author={Kamalam, GK and Joshi, Shubham and Maheshwari, Manish and Selvan, K Senthamil and Jamal, Sajjad Shaukat and Vairaprakash, S and Alhassan, Musah},
  journal={Journal of Healthcare Engineering},
  volume={2022},
  number={1},
  pages={1083978},
  year={2022},
  publisher={Wiley Online Library}
}

@inproceedings{sathyanarayana2020comparison,
  title={Comparison of head pose tracking methods for mixed-reality neuronavigation for transcranial magnetic stimulation},
  author={Sathyanarayana, Supriya and Leuze, Christoph and Hargreaves, Brian and Daniel, Bruce and Wetzstein, Gordon and Etkin, Amit and Bhati, Mahendra T and McNab, Jennifer A},
  booktitle={Medical imaging 2020: Image-guided procedures, robotic interventions, and modeling},
  volume={11315},
  pages={147--154},
  year={2020},
  organization={SPIE}
}

@article{Regmi2019BridgingTD,
  title={Bridging the Domain Gap for Ground-to-Aerial Image Matching},
  author={Krishna Regmi and Mubarak Shah},
  journal={2019 IEEE/CVF International Conference on Computer Vision (ICCV)},
  year={2019},
  pages={470-479},
  url={https://api.semanticscholar.org/CorpusID:131776514}
}

@article{Ge2024MultibranchJR,
  title={Multibranch Joint Representation Learning Based on Information Fusion Strategy for Cross-View Geo-Localization},
  author={Fawei Ge and Yunzhou Zhang and Yixiu Liu and Guiyuan Wang and Sonya A. Coleman and D. Kerr and Li Wang},
  journal={IEEE Transactions on Geoscience and Remote Sensing},
  year={2024},
  volume={62},
  pages={1-16},
  url={https://api.semanticscholar.org/CorpusID:268594513}
}

@article{Deuser2023Sample4GeoHN,
  title={Sample4Geo: Hard Negative Sampling For Cross-View Geo-Localisation},
  author={Fabian Deuser and Konrad Habel and Norbert Oswald},
  journal={2023 IEEE/CVF International Conference on Computer Vision (ICCV)},
  year={2023},
  pages={16801-16810},
  url={https://api.semanticscholar.org/CorpusID:257636648}
}

@article{Xia2024EnhancingCG,
  title={Enhancing Cross-View Geo-Localization With Domain Alignment and Scene Consistency},
  author={Panwang Xia and Yi Wan and Zhiwei Zheng and Yongjun Zhang and Jiwei Deng},
  journal={IEEE Transactions on Circuits and Systems for Video Technology},
  year={2024},
  volume={34},
  pages={13271-13281},
  url={https://api.semanticscholar.org/CorpusID:272153723}
}

@article{Zhao2021DeepLH,
  title={Deep Lucas-Kanade Homography for Multimodal Image Alignment},
  author={Yiming Zhao and Xinming Huang and Ziming Zhang},
  journal={2021 IEEE/CVF Conference on Computer Vision and Pattern Recognition (CVPR)},
  year={2021},
  pages={15945-15954},
  url={https://api.semanticscholar.org/CorpusID:233387971}
}
\bibliographystyle{icml2026}

\newpage
\appendix
\onecolumn

\newpage

\section{Appendix}\label{sec:append}
\renewcommand{\thetable}{A.\arabic{table}}
\renewcommand{\thefigure}{A.\arabic{figure}}
\setcounter{table}{0}
\setcounter{figure}{0}

\subsection{Pseudocode for the DDP Loss Computation}\label{sec:ddo-loss}

We provide the pseudocode for DDP loss computation below.

\begin{algorithm}[H]\small
\caption{Distributed InfoNCE Loss (DDP)}
\label{alg:loss}
\KwIn{Image features $\mathbf{f}_1, \mathbf{f}_2 \in \mathbb{R}^{B \times d}$, 
logit scale $\alpha$, world size $W$, rank $r$}
\KwOut{Contrastive loss $\mathcal{L}$ on rank $r$}
\BlankLine

\nl Normalize features: 
$\mathbf{f}_1 \gets \mathrm{normalize}(\mathbf{f}_1)$, 
$\mathbf{f}_2 \gets \mathrm{normalize}(\mathbf{f}_2)$\;

\nl Gather features from all processes:
\[
\{\mathbf{f}_1^{(w)}\}_{w=1}^W \gets \mathrm{AllGather}(\mathbf{f}_1), \quad
\{\mathbf{f}_2^{(w)}\}_{w=1}^W \gets \mathrm{AllGather}(\mathbf{f}_2)
\]

\nl Concatenate features, excluding local rank duplicates:
\[
\mathbf{F}_1 \gets [\mathbf{f}_1] \cup \{\mathbf{f}_1^{(w)} \mid w \neq r \}, \quad
\mathbf{F}_2 \gets [\mathbf{f}_2] \cup \{\mathbf{f}_2^{(w)} \mid w \neq r \}
\]

\nl Compute logits:
\[
\mathbf{Z}_{1} = \alpha \cdot \mathbf{F}_1 \mathbf{F}_2^\top, 
\quad
\mathbf{Z}_{2} = \mathbf{Z}_{1}^\top
\]

\nl Construct global labels:
\[
\mathbf{y} = [0, 1, \dots, N-1], \quad N = \text{rows}(\mathbf{Z}_{1})
\]

\nl Compute loss:
\[
\mathcal{L} = \tfrac{1}{2}\big( \mathrm{CE}(\mathbf{Z}_{1}, \mathbf{y}) 
+ \mathrm{CE}(\mathbf{Z}_{2}, \mathbf{y}) \big)
\]

\Return $\mathcal{L}$\;
\end{algorithm}

\subsection{Details of Human Annotation}\label{sec:human_detail}

{To ensure the accuracy of the test set, we manually revise all test samples. Our test set consists of 504 query images, each paired with both a positive and a negative reference image, resulting in 1,008 image pairs. For each pair, we generate explanations in both Chinese and English, yielding 2,016 explanations in total.}

{We engage 4 human experts in Remote Sensing, each holding at least a master's degree, to perform data annotation and revision at \$20 per hour per expert. We conduct two annotation rounds: in each round, experts verify the correctness of all four explanations (positive-Chinese, positive-English, negative-Chinese, negative-English) for each query image and correct them if necessary. If the experts cannot reach a consensus on whether a query and reference image match, we discard the query image and replace it with a suitable one from the remaining test data pool, then select new positive and negative reference images for it, followed by model-assisted generation and manual correction of the explanations. This ensures that all test samples are valid and unambiguous, providing a reliable basis for evaluation.}

{Among the 1,008 pairs (2,016 explanations), in Round 1, 596 explanations are directly modified and 68 pairs are replaced (34 query images, each with one positive and one negative pair); in Round 2, 48 explanations are directly modified and 8 pairs are replaced (4 query images, each with one positive and one negative pair). In total, 644 direct modifications are made to the explanations, and 76 pairs are replaced (38 query images, each with one positive and one negative pair) across both rounds. The decreasing number of modifications from Round 1 to Round 2 indicates progressive quality improvement and convergence toward high-quality pairs and explanations.}

\subsection{{Deployment Analysis on Edge Devices}}\label{sec:deployment}

{To evaluate the practical feasibility of our unified pipeline for real-world deployment, we conduct comprehensive performance profiling on an embedded platform. We deploy both GLEAM-C and GLEAM-X on an NVIDIA Jetson AGX Xavier (32GB RAM, JetPack 5.1) to evaluate real-world performance metrics. This analysis provides critical insights into the computational requirements and resource consumption in resource-constrained environments.}

\begin{table}[h]\tiny
\centering
\caption{{Deployment performance on Jetson Xavier (Q: query, R: references).}}
\resizebox{0.8\textwidth}{!}{%
\begin{tabular}{llcccc}
\midrule
\textbf{Module} & \textbf{Model} & \textbf{Configuration} & \textbf{Peak Power (W)} & \textbf{Peak Memory (GB)} & \textbf{Time (s)} \\ 
\midrule
GLEAM-C & ConvNeXt-B-384 & 1Q+10R & 30.4 & 5.4 & 4.3 \\ 
GLEAM-X & Qwen2.5-VL-3B & 1Q+1R & 35.7 & 12.1 & 28.5 \\ \midrule
\end{tabular}%
}
\label{tab:deploy}
\end{table}

{These measurements demonstrate that our unified pipeline can run efficiently on edge devices with moderate computational resources. All measurements are obtained under full load conditions without quantization or optimization, indicating potential for further performance improvements in production deployments.}

\subsection{{Cross-Domain Generalization Analysis}
\label{app:cross_domain}}

{We conduct an empirical cross-domain generalization experiment on ConvNeXt-B-384 to evaluate robustness across different views/modalities. We first pretrain the model on VIGOR only and test it on all four datasets without any fine-tuning. To examine whether incremental learning improves generalization, we then continue training the model on VIGOR+MAP.}

{\textbf{Scenario 1: Zero-shot Transfer (VIGOR pretrain only).} Tab.~\ref{tab:zero_shot_transfer} shows the zero-shot transfer results, which reveal significant challenges in cross-modality generalization. When trained solely on VIGOR, the model achieves 76.60\% R@1 on VIGOR but drops dramatically to 0.74\% on MAP and 0.18\% on SetVL. This substantial performance degradation stems from fundamental differences in viewpoint characteristics. These results indicate that features learned from one modality do not transfer well to drastically different view types without explicit training.}

\begin{table}[h]\tiny
\centering
\renewcommand\arraystretch{0.2} %
\caption{{Zero-shot transfer performance.}}
\resizebox{0.7\textwidth}{!}{%
\begin{tabular}{llccccc}
\midrule
\textbf{Pretrain Dataset} & \textbf{Test Dataset} & \textbf{R@1} & \textbf{R@5} & \textbf{R@10} & \textbf{Top-1} & \textbf{AP/HR} \\ \midrule
\multirow{4}{*}{VIGOR} & VIGOR & 76.60 & 95.07 & 96.83 & 99.64 & 88.29 \\ \cmidrule{2-7} 
 & University-1652 & 19.93 & 37.37 & 46.59 & 47.89 & 24.43 \\ \cmidrule{2-7} 
 & MAP & 0.74 & 2.04 & 2.74 & 11.32 & 0.74 \\ \cmidrule{2-7} 
 & SetVL & 0.18 & 0.65 & 1.12 & 4.68 & 0.18 \\ \midrule
\end{tabular}%
}
\label{tab:zero_shot_transfer}
\end{table}

{\textbf{Scenario 2: Incremental Training (VIGOR pretrain → VIGOR + MAP fine-tune).} Table~\ref{tab:incremental_training} presents the incremental training results. Performance improves only for datasets included in training: MAP performance increases significantly from 0.74\% to 93.85\% R@1, while VIGOR maintains comparable performance (77.29\% vs. 76.60\%). However, unseen modalities experience negative transfer: University-1652 drops from 19.93\% to 10.02\%, and SetVL decreases from 0.18\% to 0.04\%. This suggests that simply adding one modality does not enable generalization to other unseen view types and may even harm performance on related but unseen modalities.}

\begin{table}[h]\tiny
\centering
\caption{{Incremental training performance.}}
\renewcommand\arraystretch{0.4} %
\resizebox{0.7\textwidth}{!}{%
\begin{tabular}{llccccc}
\midrule
\textbf{Training Strategy} & \textbf{Test Dataset} & \textbf{R@1} & \textbf{R@5} & \textbf{R@10} & \textbf{Top-1} & \textbf{AP/HR} \\ \midrule
\multirow{4}{*}{VIGOR → VIGOR+MAP} & VIGOR & 77.29 & 95.25 & 96.90 & 99.66 & 88.85 \\ \cmidrule{2-7} 
 & University-1652 & 10.02 & 20.80 & 27.67 & 28.78 & 13.12 \\ \cmidrule{2-7} 
 & MAP & 93.85 & 98.35 & 98.79 & 99.88 & 93.85 \\ \cmidrule{2-7} 
 & SetVL & 0.04 & 0.24 & 0.44 & 1.92 & 0.04 \\ \midrule
\end{tabular}%
}
\label{tab:incremental_training}
\end{table}

{\textbf{Comparison with Unified Multi-Dataset Training.} In contrast to the limited generalization observed in zero-shot and incremental scenarios, our unified framework with multi-dataset training (Tab.~\ref{tab:all-c-results} in the main paper) demonstrates that when all modalities are jointly trained, the model achieves performance comparable to or better than single-dataset training. For instance, using PE-Core-L14-336 backbone, the two-phase training strategy achieves 75.96\% R@1 on VIGOR (vs. 75.53\% single-dataset), 93.97\% on MAP (vs. 92.79\%), and 23.25\% on SetVL (vs. 21.34\%). This validates that our architecture can effectively leverage diverse views and modalities when they are available during training, avoiding the negative transfer observed in incremental learning scenarios.}

{\textbf{Key Findings.} These findings highlight that while zero-shot cross-modality transfer remains challenging due to inherent viewpoint gaps, our unified approach successfully handles multiple modalities simultaneously when trained jointly. This demonstrates the importance of joint multi-dataset training for cross-view geo-localization, which is one of the primary contributions of our work.}

\subsection{{Why VIGOR as the Initial Dataset?}}
\label{app:vigor_selection}

{We choose VIGOR as the initial dataset in our two-phase training strategy based on empirical evidence from preliminary experiments. This section focuses on the ConvNeXt-B-384 model to provide a more detailed justification.}

{\noindent\textbf{Empirical justification from Tab.~\ref{tab:all-c-results}.}
As shown in Tab.~\ref{tab:all-c-results} of the main paper, we compare three training strategies: (1) single-dataset training on each dataset separately, (2) from-scratch training on merged data from all datasets, and (3) two-phase training with VIGOR pre-training followed by merged-data training. The results reveal two critical findings:}

{\textbf{1)} Direct mixed-data training degrades VIGOR performance. When training from scratch on merged data, VIGOR Recall@1 drops to 73.37\% compared to 76.60\% in single-dataset training. This suggests that the model struggles to learn basic CVGL capabilities when simultaneously handling multiple datasets with different characteristics.}

{\textbf{2)} VIGOR pre-training enables effective multi-dataset learning. Using VIGOR as the initial phase, the two-phase strategy achieves 75.66\% Recall@1 on VIGOR (recovering most of the performance loss) while significantly improving on other datasets: MAP increases from 92.52\% to 94.05\%, and SetVL-480K improves from 14.33\% to 15.28\%.}

{\noindent\textbf{Comparison with other pre-training datasets.}
To validate the choice of VIGOR, we conduct additional experiments on ConvNeXt-B-384 to compare different initial datasets (VIGOR, University-1652, and SetVL-480K) before mixed-data training. Results are shown in Tab.~\ref{tab:pretrain_comparison}.}

\begin{table}[h]\tiny
\centering
\renewcommand\arraystretch{0.8} %
\caption{{Comparison of different initial datasets for two-phase training strategy.}}
\resizebox{0.8\textwidth}{!}{%
\begin{tabular}{lcccc}
\midrule
\textbf{Training Strategy} & \textbf{VIGOR R@1} & \textbf{University R@1} & \textbf{MAP R@1} & \textbf{SetVL R@1} \\ \midrule
\textbf{VIGOR → Merge (Tab. 2)} & 75.66 & 87.03 & 94.05 & 15.28 \\ %
\textbf{University → Merge} & 73.19 & 81.52 & 92.32 & 14.30 \\ %
\textbf{SetVL → Merge} & 70.55 & 75.64 & 91.93 & 14.39 \\ \midrule
\end{tabular}%
}
\label{tab:pretrain_comparison}
\vspace{-1em}
\end{table}

{Three key observations emerge from Tab.~\ref{tab:pretrain_comparison}:}

{\textbf{1)} VIGOR-first significantly outperforms all alternatives. It achieves +2.47\% on VIGOR, +5.51\% on University-1652, +1.73\% on MAP, and +0.98\% on SetVL compared to University-first pre-training, demonstrating its superior ability to provide effective initialization for multi-dataset learning.}

{\textbf{2)} University-first shows moderate but suboptimal performance. While it provides reasonable initialization, it significantly underperforms VIGOR-first, particularly on University-1652 itself (81.52\% vs. 87.03\%), suggesting it provides weaker cross-dataset transferability.}

{\textbf{3)} SetVL-first performs worst across all datasets. Despite being the largest dataset, SetVL's extreme difficulty (R@1 $\sim$14\%) and highly diverse scene types result in poor performance on VIGOR (70.55\%) and University-1652 (75.64\%), making it unsuitable for initial feature learning.}

{\noindent\textbf{Why does VIGOR work better?}
VIGOR provides superior initialization due to its \textit{balanced difficulty}. As shown in Tab.~\ref{tab:all-c-results}, VIGOR achieves moderate performance in single-dataset training (R@1 76.60\%), indicating it is neither too easy (like MAP at 92.60\%) nor too difficult (like SetVL at 14.23\%). This balanced difficulty allows the model to learn fundamental cross-view correspondence reasoning without overfitting to trivial patterns or failing to converge on overly challenging scenarios. Consequently, VIGOR pre-training establishes a robust feature space that facilitates effective adaptation to diverse datasets in the subsequent mixed-data training phase.}

\subsection{{More Semantic Evaluation Methods}}\label{sec:more_judge}

{
Beyond Sentence-BERT, we introduce two additional evaluation methods to comprehensively assess the quality of generated explanations in the EN setting.}

{
First, we conduct an LLM-as-a-judge evaluation using Gemini 2.5 Flash with a detailed 0-5 scoring rubric focusing on identifying geographical features and spatial reasoning accuracy. We choose Gemini 2.5 Flash because it is independent of the training data sources and less likely to show preference bias. Importantly, we provide Gemini with the query image, reference image, reference explanation, and model-generated explanation, enabling it to assess the factual correctness of the reasoning rather than just linguistic quality. Second, we perform a human evaluation using the same rubric with 4 human experts in Remote Sensing. To improve consistency in human evaluation, each expert evaluates one complete dataset among MAP, SetVL-480K, University-1652, and VIGOR. The detailed scoring rubric is provided in the Sec.~\ref{sec:eval_prompt}.}

\begin{table}[h]\tiny
\centering
\renewcommand\arraystretch{0.7} %
\caption{{LLM-as-a-judge evaluation results with  Gemini 2.5 Flash.}}
\resizebox{0.75\textwidth}{!}{%
\begin{tabular}{lccccccc}
\midrule
\textbf{Gemini   2.5 Flash} & \textbf{} & \textbf{} & \textbf{} & \textbf{} & \textbf{} & \textbf{} & \textbf{} \\ \midrule
\textbf{Model} & \textbf{0} & \textbf{1} & \textbf{2} & \textbf{3} & \textbf{4} & \textbf{5} & \textbf{Avg Score} \\ \midrule
\textbf{GPT-4o} & 17.5\% & 0.2\% & 1.6\% & 7.5\% & 65.2\% & 8.0\% & 3.27 \\ %
\textbf{Doubao-1.5} & 19.5\% & 0.0\% & 1.0\% & 4.1\% & 66.0\% & 9.4\% & 3.25 \\ \midrule
\textbf{Qwen2.5-VL-3B-Instruct} & 42.2\% & 2.3\% & 17.0\% & 14.3\% & 23.2\% & 1.1\% & 1.77 \\ %
\textbf{\quad+label only} & 90.7\% & 0.2\% & 0.5\% & 0.8\% & 7.4\% & 0.4\% & 0.35 \\ %
\textbf{\quad+GPT-4o} & 8.7\% & 0.2\% & 3.0\% & 6.6\% & 65.2\% & 16.3\% & 3.68 \\ %
\textbf{\quad+Doubao-1.5} & 8.6\% & 0.1\% & 1.9\% & 10.9\% & 63.7\% & 14.8\% & 3.65 \\ \midrule
\end{tabular}%
}
\label{tab:gemini_eval}
\end{table}

\begin{table}[h]\tiny
\centering
\caption{{Human evaluation results with 4 experts.}}
\renewcommand\arraystretch{0.8} %
\resizebox{0.75\textwidth}{!}{%
\begin{tabular}{lccccccc}
\midrule
\textbf{Human Test} & \textbf{} & \textbf{} & \textbf{} & \textbf{} & \textbf{} & \textbf{} & \textbf{} \\ \midrule
\textbf{Model} & \textbf{0} & \textbf{1} & \textbf{2} & \textbf{3} & \textbf{4} & \textbf{5} & \textbf{Avg Score} \\ \midrule
\textbf{GPT-4o} & 11.9\% & 9.4\% & 20.8\% & 15.4\% & 36.2\% & 6.3\% & 2.73 \\ %
\textbf{Doubao-1.5} & 13.4\% & 10.4\% & 15.7\% & 12.4\% & 40.2\% & 7.9\% & 2.79 \\ \midrule
\textbf{Qwen2.5-VL-3B-Instruct} & 28.3\% & 22.3\% & 24.0\% & 9.4\% & 14.1\% & 1.9\% & 1.64 \\ %
\textbf{\quad+label only} & 66.5\% & 13.4\% & 14.4\% & 2.5\% & 2.8\% & 0.5\% & 0.63 \\ %
\textbf{\quad+GPT-4o} & 6.6\% & 8.1\% & 19.1\% & 13.7\% & 41.5\% & 11.1\% & 3.09 \\ %
\textbf{\quad+Doubao-1.5} & 5.8\% & 6.9\% & 20.6\% & 15.4\% & 39.5\% & 11.8\% & 3.11 \\ \midrule
\end{tabular}%
}
\label{tab:human_eval}
\end{table}

{
The results are shown in Tab.~\ref{tab:gemini_eval} and Tab.~\ref{tab:human_eval}. We observe consistent patterns across all three metrics. In the LLM-as-a-judge evaluation, the explanation-supervised models achieve average scores of 3.68 and 3.65, even surpassing their teacher models GPT-4o (3.27) and Doubao-1.5 (3.25), while substantially outperforming the baseline Qwen2.5-VL-3B-Instruct at 1.77. The human evaluation confirms this trend with scores of 3.09 and 3.11 versus 2.73 and 2.79 for the teachers and 1.64 for the baseline. The label-only model scores near zero (0.35 and 0.63) as it cannot generate explanations. These results align well with the Sentence-BERT scores in Tab.~\ref{tab:x-sim} of the main text, where the explanation-supervised models show similar improvements. The consistency across semantic similarity, LLM judgment, and human assessment demonstrates that the explanation-supervised models produce high-quality reasoning.
}

\subsection{{Potential Biases in Data Curation}\label{sec:main_dis_x}}

{GLEAM-X relies on GPT-4o and Doubao-1.5 to generate explanation labels, which may introduce biases such as linguistic patterns or superficial reasoning into the training data, raising concerns about propagating teacher model limitations to student models.}

{Our approach mitigates these concerns through ground-truth guidance during data curation. When generating explanations, we provide GT matching labels to teacher models, steering them toward correct reasoning directions. As shown in Tab.~\ref{tab:x-acc} and Tab.~\ref{tab:x-sim}, teacher models (GPT-4o and Doubao-1.5) achieve 77-84\% average accuracy and 0.74-0.83 average similarity when evaluated without GT labels during inference. In contrast, student models (Qwen2.5-VL-3B) trained on GT-guided explanations achieve 87-91\% accuracy and 0.79-0.84 similarity, demonstrating that GT supervision effectively controls bias and enables students to even surpass teacher performance.}

The quality of generated explanations depends on the teacher model's capabilities, which determine the ceiling for knowledge transfer. In this work, we have not performed additional data refinement (e.g., human filtering or iterative improvement). Nevertheless, the strong performance indicates that the current explanation quality is sufficient for effective knowledge transfer. As this is the first work to introduce explanation supervision for cross-view geo-localization, improving data quality through advanced curation strategies remains a promising direction for future work.

\newpage

\subsection{Hyper-parameters}

\subsubsection{GLEAM-C}

Here we provide the training hyper-parameters of GLEAM-C using PE-Core-L14-336 ViT on the second training phase.

\begin{table}[h]
\centering
\caption{Training hyper-parameters of GLEAM-C using PE-Core-L14-336.}
\label{tab:hyperparameters}
\begin{tabular}{lc}
\midrule
\textbf{Parameter} & \textbf{Value} \\
\midrule
Model & PE-Core-L14-336 \\
Epochs & 40 \\
Batch Size (Total) & 300 \\
Learning Rate & 1e-4 \\
Scheduler & Cosine \\
Warmup Epochs & 1 \\
Label Smoothing & 0.1 \\
Neighbour Select & 64 \\
Neighbour Range & 128 \\
Prob Rotate & 0.75 \\
Prob Flip & 0.5 \\
\midrule
\end{tabular}
\end{table}

\subsubsection{GLEAM-X}

Here we provide the hyper-parameters of training Qwen2.5-VL-3B-Instruct on GLEAM-X.

\begin{table}[ht]
\centering
\caption{Hyper-parameters of training Qwen2.5-VL-3B-Instruct on GLEAM-X.}
\begin{tabular}{lc}
\midrule
\textbf{Parameter} & \textbf{Value} \\
\midrule
Learning Rate & 5e-5 \\
Batch Size (Total) & 60 \\
Epochs & 1 \\
Weight Decay & 0.1 \\
Warmup Ratio & 0.05 \\
Max Length & 2048 \\
Min/Max Pixels & 256/1296 \\
Gradient Checkpointing & True \\
Precision & BF16 \\
\midrule
\end{tabular}
\end{table}

\newpage

\subsection{More Data Examples of GLEAM-X.}

\begin{figure}[h]
    \centering
    \includegraphics[width=0.9\linewidth]{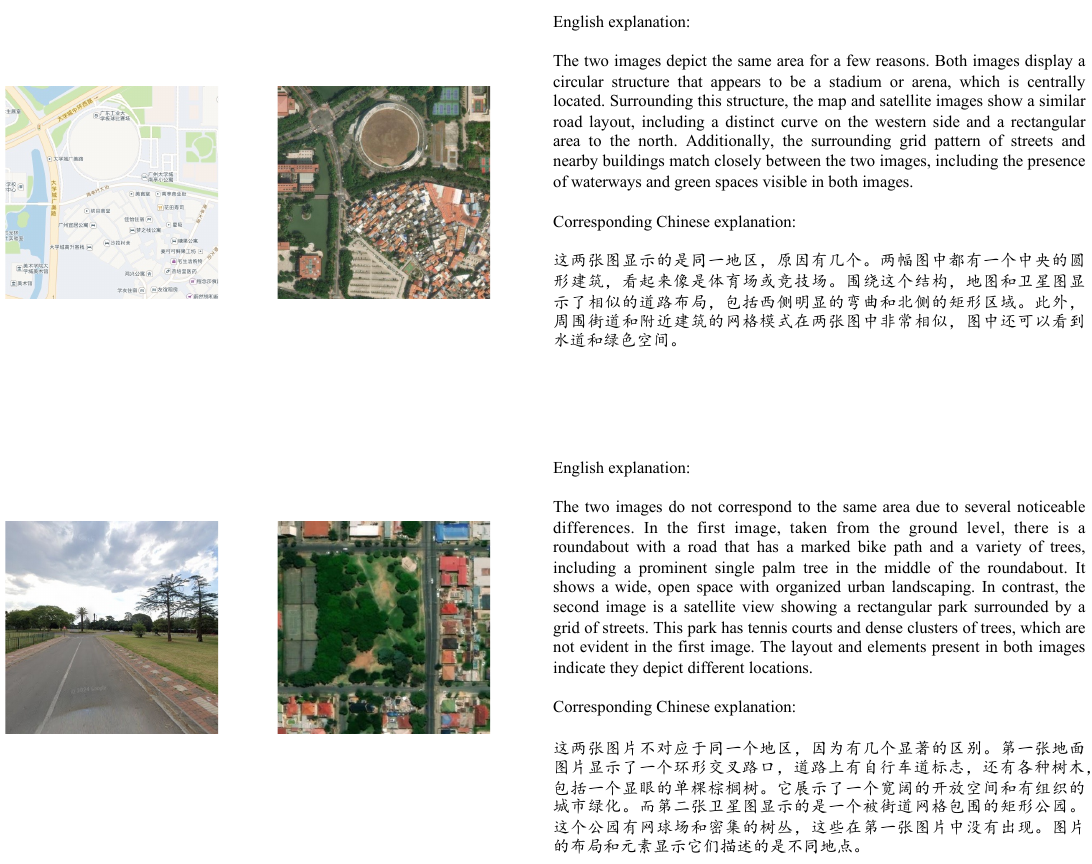}
    \caption{Street map and ground photograph samples.}
    \label{fig:sample1}
\end{figure}

\begin{figure}[t]
    \centering
    \includegraphics[width=0.9\linewidth]{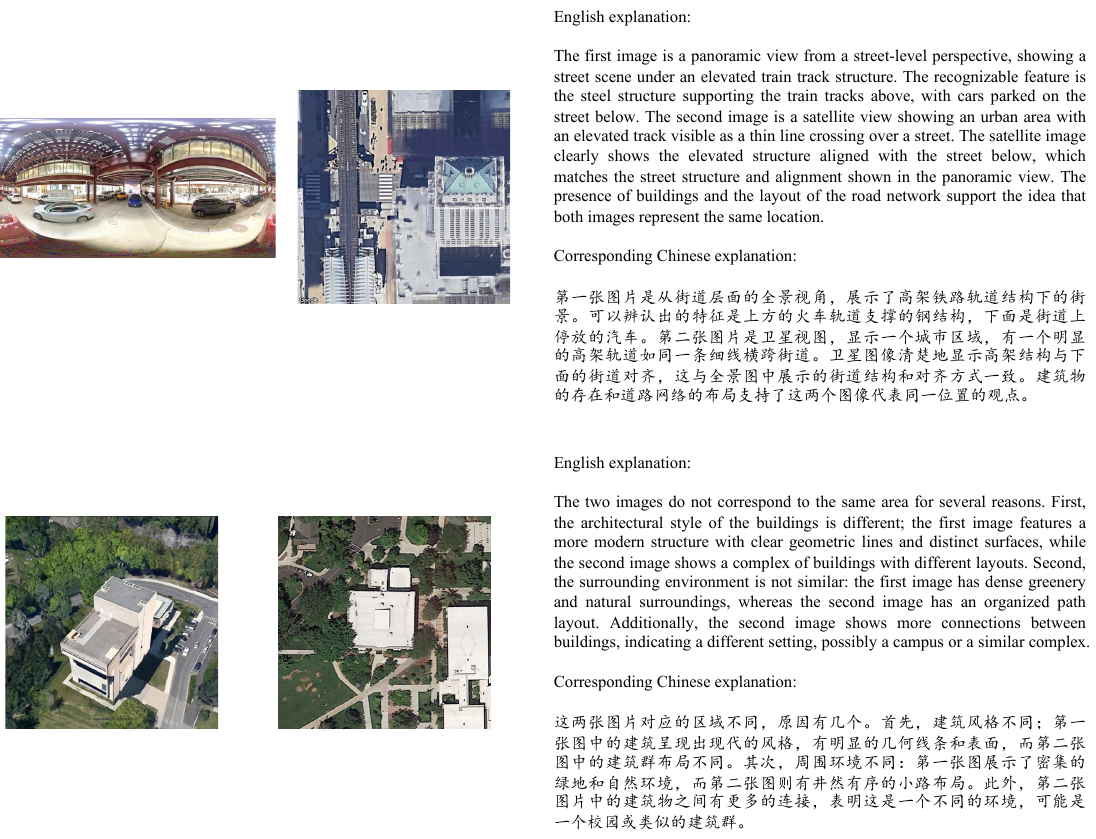}
    \caption{Panoramic view and UAV imagery samples.}
    \label{fig:sample2}
\end{figure}

\newpage

\subsection{More Qualitative Results on GLEAM-X Benchmark}

\begin{figure}[t]
    \centering
    \vspace{-3em}
    \includegraphics[width=0.8\linewidth]{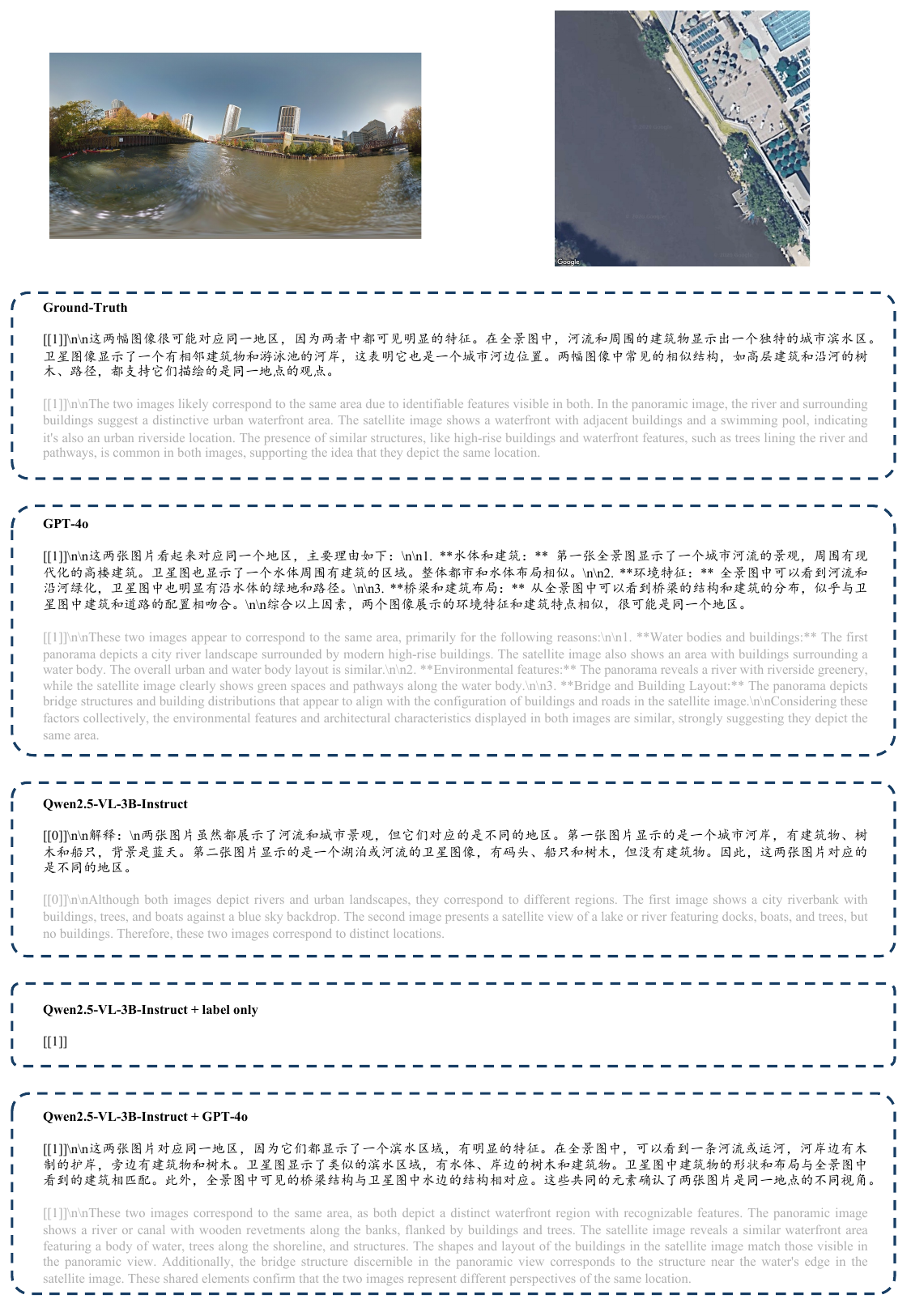}
    \caption{Sample on the VIGOR test set (Chinese scenario). The gray English text is a direct translation of the Chinese response.}
    \vspace{-3em}
\end{figure}

\begin{figure}[t]
    \centering
    \vspace{-1em}
    \includegraphics[width=0.8\linewidth]{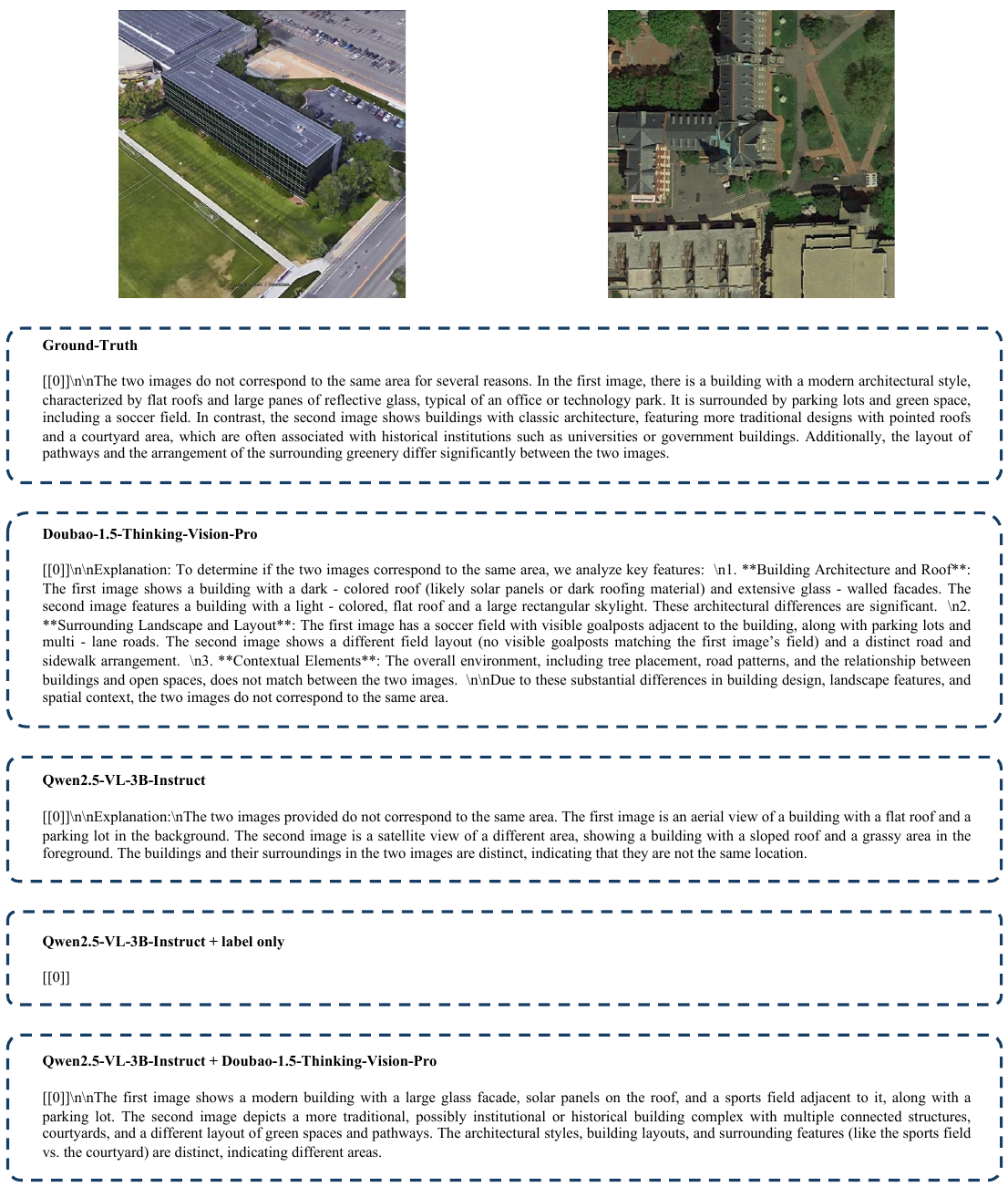}
    \caption{Sample on the University-1652 test set (English scenario).}
\end{figure}

\clearpage

\subsection{{Evaluation Rubric for LLM-as-a-Judge and Human Evaluation}}
\label{sec:eval_prompt}
{
\begin{lstlisting}
judge_en_prompt = '''
# Task Introduction

This task aims to evaluate the quality of explanations for cross-view geo-localization. Given two images (which may include satellite imagery, UAV imagery, street maps, panoramic images, and ground photos), annotators must determine whether they depict the same geographical location and provide a detailed explanation for their judgment. The explanation should identify specific geographical features and demonstrate spatial reasoning to support the matching decision. You are now provided with two images to evaluate, a model answer, and a reference answer. Please score the model answer strictly according to the following scoring criteria (0-5 points) and provide justification for your score.

# Scoring Criteria

## 5 Points - Excellent
The explanation quality significantly exceeds the reference annotation, identifying more correct specific geographic features, or demonstrating markedly superior precision and depth in spatial reasoning compared to the reference annotation. The argumentation is logically rigorous, feature descriptions are highly verifiable and distinctive, with no factual errors. Demonstrates exceptional spatial analysis capabilities and can provide equally valid argumentative dimensions not covered by the reference annotation.

Example: [[1]]\n\nBoth images show the same area in Munich: An oval-shaped stadium is visible in the center of the satellite image, labeled as Olympiastadion on the map; the curved lake shape on the west side of the stadium matches perfectly; the spacing and orientation of three parallel roads on the north side are consistent; the rectangular parking lot position in the southeast corner corresponds; additionally, a characteristic spiral ramp structure is visible on the south side of the stadium, which precisely matches the building outline on the map.

## 4 Points - Good (Reference annotations typically correspond to 4-point level)
The explanation identifies geographic features that are essentially consistent with the reference annotation or other correct related geographic features, with accurate spatial reasoning, coherent logic, and no critical errors. Can adequately explain the main basis for match or non-match, reaching the quality level of the reference annotation. Directional descriptions or feature positioning may be somewhat general, but this does not affect the validity of the argumentation.

Example: [[0]]\n\nThe two images do not correspond: The map shows a regular grid-pattern block with obvious roundabouts and radial roads, while the satellite image shows irregular curved roads surrounded by large areas of farmland. The road patterns and land use types are clearly different.

## 3 Points - Acceptable
The explanation identifies some correct features but lacks specificity in description, spatial analysis is relatively superficial, and argumentation is insufficient. Although the conclusion is correct, the reasoning process has vagueness or generalization issues. There may be minor factual errors that do not affect the final judgment, such as slight misjudgment of feature types.

Example: [[1]]\n\nThe two images match because both show rivers and bridges. The water body shape in the satellite image is consistent with the river seen in the panoramic image, and both are surrounded by buildings.

## 2 Points - Poor
Although the match judgment is correct, the explanation contains critical factual errors, spatial logic fallacies, or identified features are overly generalized and lack distinctiveness. For non-match cases, may misjudge image modality differences (perspective, color, resolution) as geographic feature differences. The reasoning lacks validity and is difficult to provide substantial support for the judgment.

Example: [[0]]\n\nNo match, because the buildings in the first image have red roofs, while the building roofs in the second satellite image are gray. The different colors indicate different places.

## 1 Point - Very Poor
Although the match judgment is correct, the reasoning process has fundamental flaws, including hallucinating non-existent features, logical confusion, or using task-irrelevant factors (image quality, shooting conditions, weather conditions) as primary arguments. Demonstrates serious misunderstanding of the task objective.

Example: [[0]]\n\nDoes not correspond, because the first is a panoramic image taken during the day, and the second is a satellite image, with different shooting methods. Moreover, the first image is clearer, while the second is more blurry.

## 0 Points - Fail
The match judgment is incorrect, or no valid explanation is provided (blank, label only, off-topic).

---

# Evaluation Dimensions

**Feature Accuracy:** Whether the identified geographic features actually exist. Can be consistent with the reference annotation or provide other valid correct features. Avoid feature hallucination or misidentification.

**Reasoning Validity:** Whether the argumentation adequately supports the match judgment. For match cases, need to demonstrate consistency of spatial features; for non-match cases, need to demonstrate essential differences in geographic features rather than image modality differences.

**Description Specificity:** Whether feature descriptions are distinctive and localizable. High-quality explanations should include specific spatial positioning information, rather than generalized descriptions applicable to any location.

**Verifiability:** All statements should be verifiable through image observation, avoiding subjective speculation or unverifiable assertions.

---

# Output Format (Must Strictly Follow)

Format: [[<score>]]\n\n<brief explanation>

- <score>: Integer between 0-5
- <brief explanation>: One sentence explaining the scoring rationale

**Output Examples:**

[[5]]\n\nIdentified 7 precisely matching geographic features with accurate spatial positioning and rigorous argumentation.

[[3]]\n\nConclusion is correct but feature descriptions are too general, lacking specific spatial positioning information.

[[0]]\n\nMatch judgment is incorrect.
'''
\end{lstlisting}

}

\end{document}